\documentclass{article}

\usepackage{arxiv}
\usepackage[utf8]{inputenc} 
\usepackage[T1]{fontenc}    
\usepackage{hyperref}       
\usepackage{url}            
\usepackage{booktabs}       
\usepackage{amsfonts}       
\usepackage{nicefrac}       
\usepackage{microtype}      
\usepackage{lipsum}

\usepackage{url} 
\usepackage[ruled,lined]{algorithm2e}
\usepackage{graphicx}
\usepackage{multirow}
\usepackage{mymacros}
\usepackage{amsmath}
\usepackage{float}
\usepackage{listings}
\usepackage{amsmath}
\usepackage{float}
\usepackage{longtable}
\usepackage{subcaption}

\title{Human Face Expressions from Images\\ 
{\small 2D Face Geometry and 3D Face Local Motion versus Deep Neural Features}}

\author{
  Rafa\l\ Pilarczyk \\
  \texttt{rpilarcz@gmail.com} \\
   \And
  Xin Chang\\
   \And
  W\l adys\l aw Skarbek\\
  \texttt{w.skarbek@ire.pw.edu.pl} \\
   \And
   Affiliation\\
    Institute of Radioelectronics and Multimedia Technology \\
    Faculty of Electronics and Information Technology \\ 
    Warsaw University of Technology \\ 
    Nowowiejska 15/19, 00-665 Warszawa, Poland\\
}

\begin{document}

\maketitle

\begin{abstract} 
Several computer algorithms for recognition of visible human emotions are compared at the web camera scenario using CNN/MMOD face detector. The recognition refers to four face expressions: smile, surprise, anger, and neutral. At the feature extraction stage, the following three concepts of face description are confronted: (a) static 2D face geometry represented by its 68 characteristic landmarks (FP68); (b) dynamic 3D geometry defined by motion parameters for eight distinguished face parts (denoted as AU8) of personalized Candide-3 model; (c) static 2D visual description as 2D array of gray scale pixels (known as facial raw image). At the classification stage, the performance of two major models are analyzed: (a) support vector machine (SVM) with kernel options; (b) convolutional neural network (CNN) with variety of relevant tensor processing layers and blocks of them. The models are trained for frontal views of human faces while they are tested for arbitrary head poses. For geometric features, the success rate (accuracy)  indicate  nearly triple increase of performance of CNN with respect to SVM classifiers. For raw images, CNN outperforms in accuracy  its best geometric counterpart (AU/CNN) by about 30 percent while the best SVM solutions are inferior nearly four times. For F-score the high advantage of raw/CNN over geometric/CNN and geometric/SVM is observed, as well. We conclude that contrary to CNN based emotion classifiers, the generalization capability wrt human head pose is for SVM based emotion classifiers poor.  
\end{abstract}

\keywords{ expression recognition \and face landmarks \and  facial action units\and  SVM classifier\and  CNN classifier
}


\section{Introduction}

Recognition of human face expression is an useful functionality in computer applications based on human-computer-Interfacing (HCI). The algorithmic background for such systems belongs to artificial intelligence (AI) in general with strong share of computer vision research and development.

Facial expressions as a natural non-verbal means for communication, are conveying human feelings. This language was developed in human beings evolution as a irreplaceable tool in their mutual communication.

Research of the passed  centuries including recent findings of psychologists and physiologists,  suggest that the visible emotions,  regardless of the sexual, national, or religious aspects can be classified into the six basic forms (Ekman and Friesen  \cite{Ekman1971ConstantsAC}): {\it happiness, anger, fear, disgust, surprise, and sadness.} In recognition systems, for the completeness, the {\it neutral} category is appended.

%

The automatic recognition of facial expression attracted many researchers in the recent $30$ years of image analysis development. For instance, already in year 1998 the authors from famous Pittsburgh's group (Lien, Kanade, Cohn, Li  \cite{670980})  compared four types of facial expressions getting the accuracy of about $90\%$ for the frontal pose. However, the algorithm recognizes only upper face expressions in the forehead and brow regions. Interestingly, the facial expressions are represented by combinations of action units defined $20$ years earlier as the Facial Action Coding System (FACS) by Ekman and Friesen in  \cite{ekman:friesen:1978}. The method uses Hidden Markov Model (HMM) to recognize "words" of action units.

The FACS concept had significant impact on  graphics standardization  conducted within MPEG-4 group on 3D model of human head. The works had led to Candide-3 model  \cite{Ahlberg01candide-3} including numerical values for model vertexes and mesh indexes for model standard geometry, shape units for geometry individualization, and action units for model animation. Introducing Candide model happened to be the important step in image face analysis via 3D models. Then (year 2001), the missing factor was a tool for detecting facial landmarks in the fast and the accurate way. There was general opinion that such task is beyond of the contemporary computing technology. 

The image analysis research was facing then the mythical "curse of high dimensionality modeling" and the optical/pixel flow tools were the evidence of the crisis.  

However, after a decade both the computing technology and algorithms have made the enormous progress and getting in real time the 3D models of real human heads becomes the reality in human-computer interfaces HCI (as shown for instance in \cite{ycs-smile}). Nearly, in the same time the "neural revolution" arrived in the form of deep neural networks (DNN) changing dramatically the status of "intelligent applications" including visible emotions recognition in real time. 

Deep learning approaches, particularly using convolutional neural networks (CNN) have been successful at computer vision related task due to achieving good generalization from multidimensional data \cite{imagenet2,imagenet,vgg-network,inception-neural}. Facial expression and emotion recognition with deep learning methods were used in \cite{liu_emo_wild,sikka_emo_wild,liu_facial_emo,kanou_neural_emo,tang_svm_neural_emo,expression-image-neural}. Moreover DNN network were jointly learned with SVM by Tang\cite{tang_svm_neural_emo} for emotion classification. There is great amount of databases for facial expression recognition from image as well as sophisticated recognition methods used in the wild. 

\section{Geometric feature extraction}

\begin{figure}[htb]
   \begin{center}
   \begin{tabular}{cc}
   \includegraphics[width=.475\linewidth]{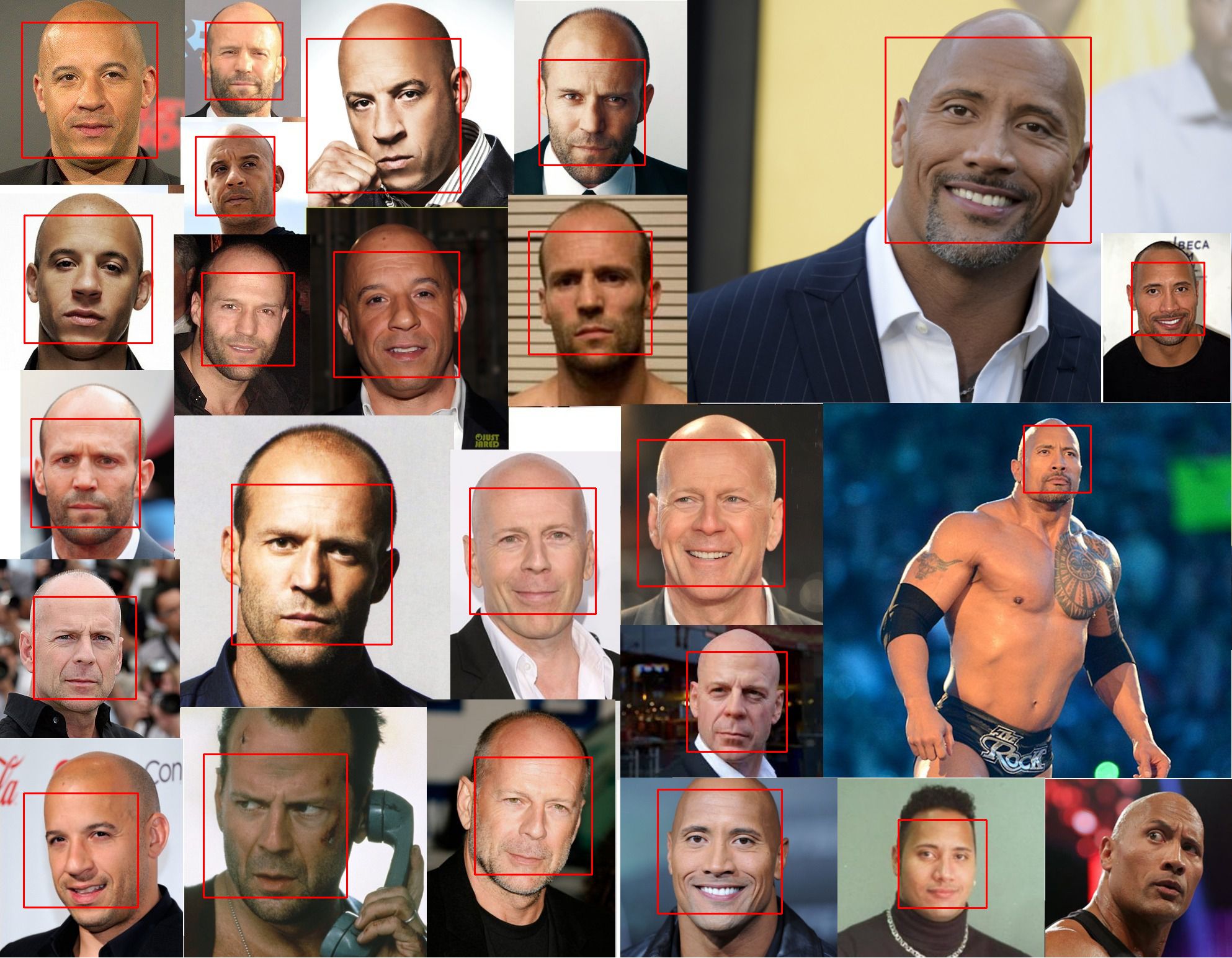} 
   &
   \includegraphics[width=.475\linewidth]{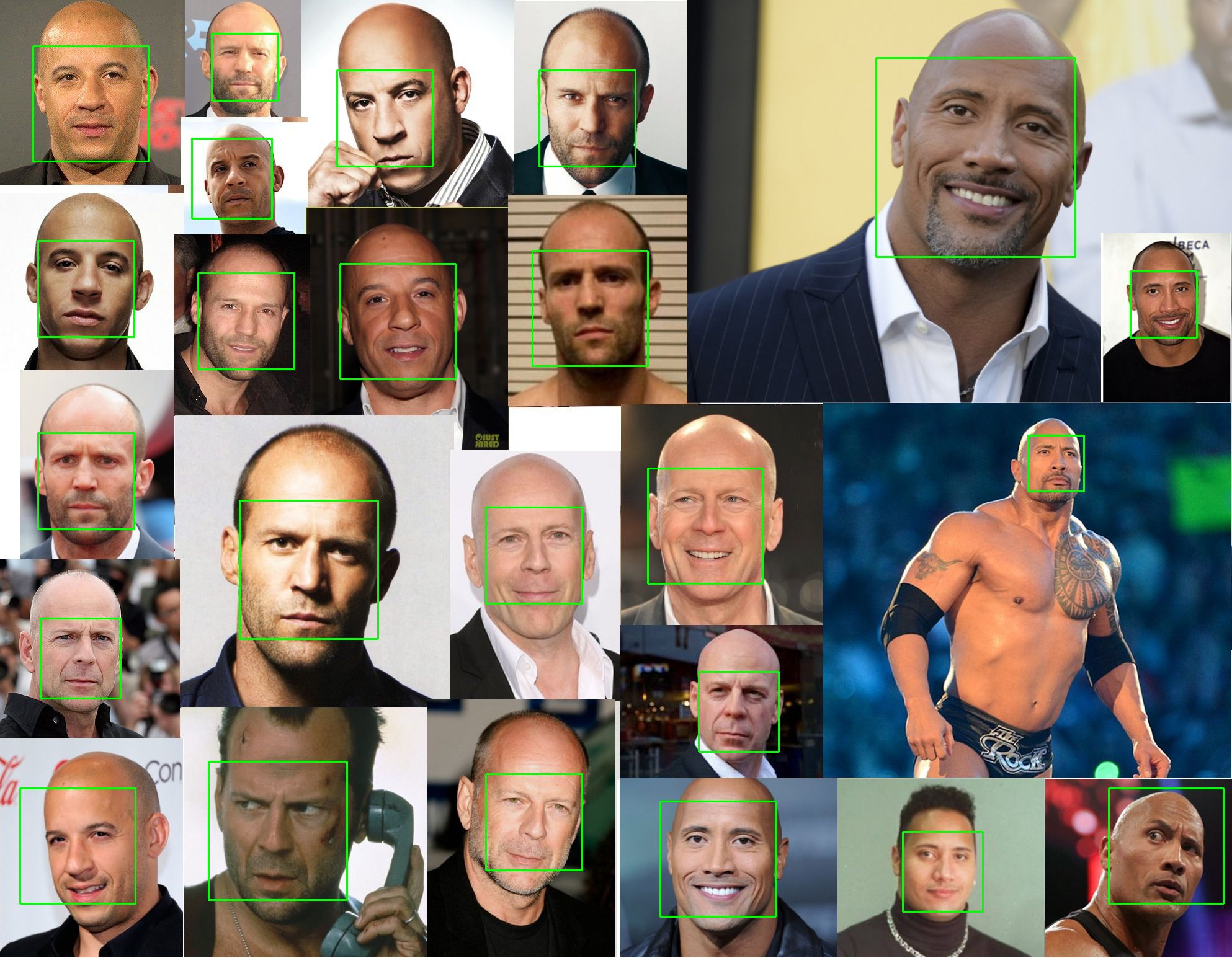} 
   \end{tabular}
   \end{center}
   \caption[example] 
{\label{fig:viola-versus-king}
Face detectors: (left) Viola and Jones algorithm ({\tt opencv}); (right) King's method ({\tt dlib}).
}
\end{figure}

On top of all facial image analysis usually the face detector is exploited. In our research we use the face detector of King  \cite{king-15} available in {\tt dlib} library  \cite{king-dlib}. It is a novel proposal which appears to be more accurate than Viola and Jones face detector  \cite{viola-jones} (cf. Fig.\ref{fig:viola-versus-king}) available in {\tt opencv} library and being still improved by retraining for new data sets.

The face detector offered in {\tt dlib} is an instance of more general object detector which can be tuned for various object classes and various feature extractors. 

The generality of the maximum margin object detector (MMOD) in {\tt dlib }is based on similar concepts to those which were proposed by Joachims et alters \cite{joach-09b} at developing the ideas of Structural Support Vector Machine (SSVM). 

Comparing to Viola and Jones algorithm for face detection, instead of the boosted cascade of weak classifiers based on  ensembles of local region contrasts, we have in Kings proposals both, the HOG features (HOG/MMOD), i.e. the Histogram of Oriented Gradients  \cite{dalal-triggs}, and features extracted by trained neural network (CNN/MMOD), as well. The SVM model is trained by max-margin convex optimization problem defined for collections of image rectangles.

The sophisticated method of selection of image rectangles with objects of interest avoids the combinatorial explosion by SSVM  trick where only the worst constraints are taken to relevant quadratic optimization and by a smart heuristics which in the greedy way allows us to get suboptimal rectangle configurations for a complex but convex risk function.

\subsection{Facial characteristic landmarks}

At the feature extraction stage, the following three concepts of face description are confronted: (a) static 2D face geometry represented by its 68 characteristic landmarks (FP68); (b) dynamic 3D geometry defined by motion parameters for eight distinguished face parts (denoted as AU8) of personalized Candide-3 model; (c) static 2D visual description as 2D array of gray scale pixels (known as facial raw image).

\begin{figure}[htb]
\begin{tabular}{cc}
\begin{tabular}{r}
    \includegraphics[width=0.5\textwidth]{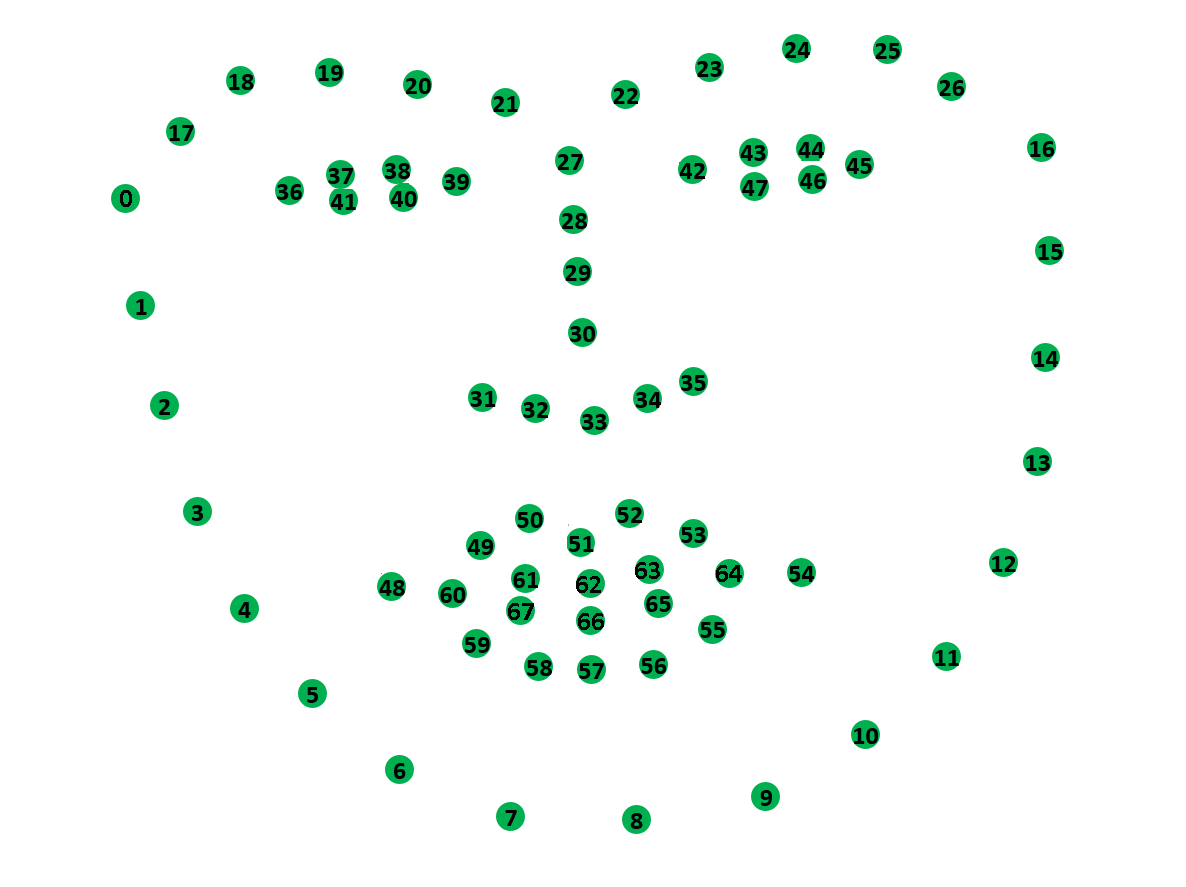}
\end{tabular}
&
\begin{tabular}{l}
\includegraphics[width=.25\linewidth]{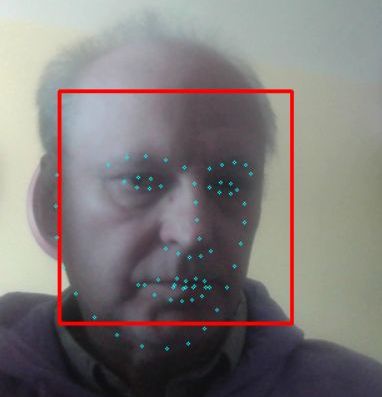}
\\
\includegraphics[width=.25\linewidth]{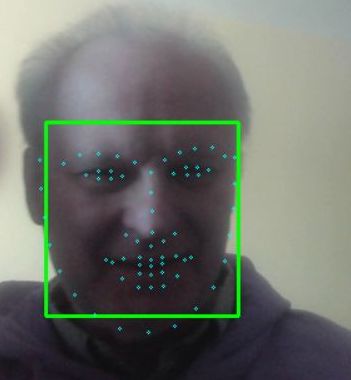}
\end{tabular}
\end{tabular}
    \caption{\label{fig:detect-ws}FP68 indexing (left) and detection boxes with FP68 points on faces (right).}
\end{figure}

The FP68 detector is implemented as well  in {\tt dlib}  \cite{king-dlib} library as well. The detector exploits many regression trees for 68 HOG features mapping. The cascade approach using of many small regression trees gives more effective detector than using
one large regression model. The trees are built using stochastic gradient boosting of Friedman  \cite{scikit-learn}.

On the other hand {\tt dlib} FP68 detector is unsatisfactory for detecting landmarks on non-frontal face images. In most cases landmarks are incorrectly marked especially for test database. We switch to CNN-based FP68 regressor \cite{landmarks-pilarczyk-skarbek} built at the top of MobileNet-V1 \cite{mobilenets-2017} which is trained on public datasets 300W, 300VW, IBUG \cite{300w-database,300vw-first,300w-semiauto,300vw-offlinedeform,300VW} in order to extract missing facial landmarks.

\subsection{Candide-3 model personalization and animation}
   
\begin{figure}[htb]
\hspace*{-5mm}
\begin{tabular}{ccc}
\begin{tabular}{r}
\includegraphics[width=0.33\linewidth]{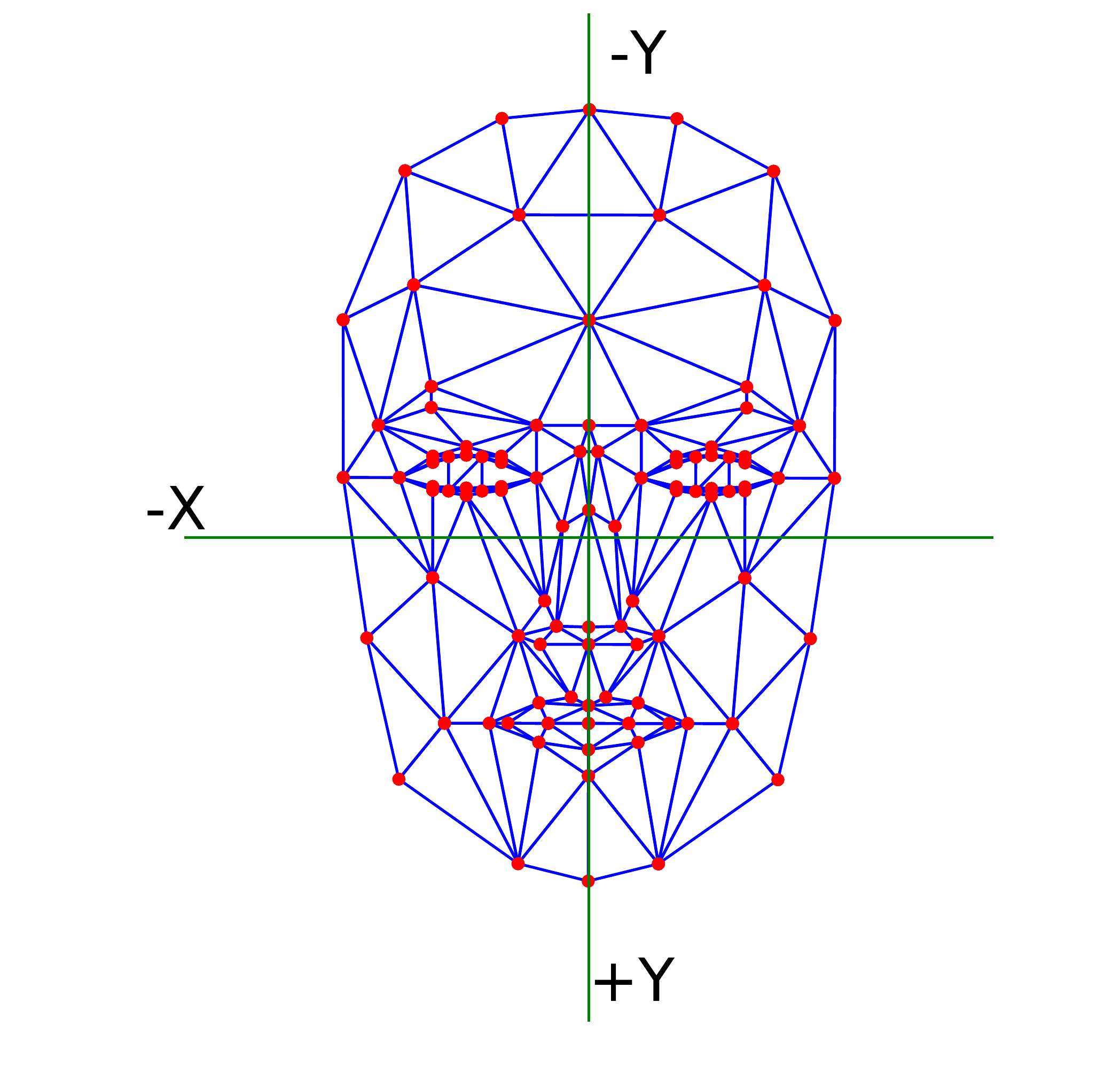}
\end{tabular}
&
\begin{tabular}{l}
    \includegraphics[width=0.18\textwidth]{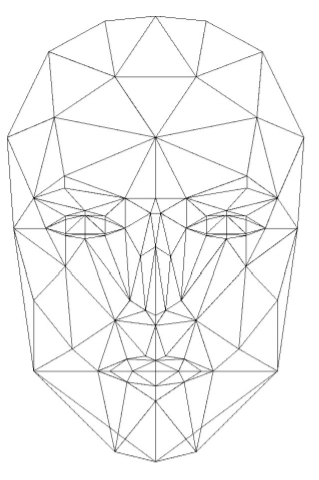}
\end{tabular}
&
   \begin{tabular}{c}
   \includegraphics[width=.33\linewidth]{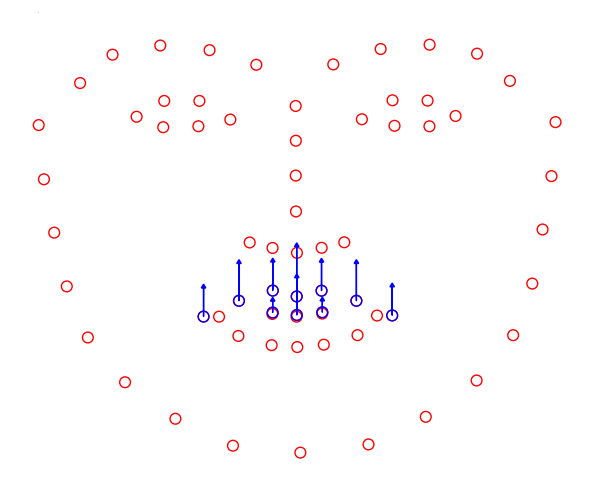} 
\\
   \includegraphics[width=.33\linewidth]{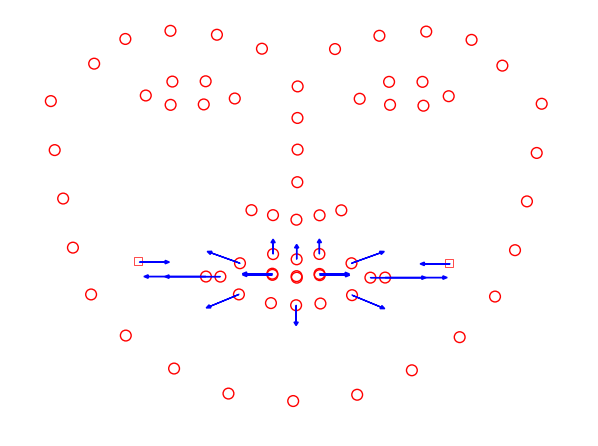} 
   \end{tabular}
\end{tabular}    
\caption{\label{fig:Candide-3}Candide-3 wire-frame model (XY view), the 3D model with the removed hidden invisible lines (with some triangles removed), and the motion of two action units (AU10, AU20) to be identified via 3D model controlled by FP68 points.}
\end{figure}

   Candide-3 model (cf. Fig.\ref{fig:Candide-3}) consists of static human head geometry specified by a discrete set of 3D points (also known as mean Candide shape) with coordinates normalized to the spatial cube $[-1,+1]^3$ and the dynamic geometry specified via motion vectors assigned to distinguished groups of 3D points (also known as shape units and action units).

\subsubsection{Concept of action and shape units}   
   The Candide-3 model can be scaled independently in shape and action units, i.e. a local motion can be performed. While the shape units can be used to personalize the face features (e.g. mouth width) the dynamic scaling of action units enables simulation of local face motions typical for face expressions \cite{au-chang-skarbek}.

After local motions, global rotation is applied followed by global scaling and global translation. 
While local motions (personalization and animation) are performed in Cartesian coordinate system of Candide-3, the global scaling and translation are usually used to transfer the model to an observer coordinate system. If the observer is a web camera then the face image can be used to identify parameters of local motions provided we have a fast algorithm identifying in the image the points which convey the local motion. 

In the identification of local motion parameters we have to know also the global affine motion parameters (rotation, scaling, and translation). To identify the global motion we need also points of model which do not belong to groups of shape and action units. Fortunately, we can select 37 facial salient points out of FP68 landmarks to be used to minimize the modeling error.

\begin{figure}[htb]
\hspace*{-5mm}
 \begin{tabular}{cccc}
    \includegraphics[width=0.22\textwidth]{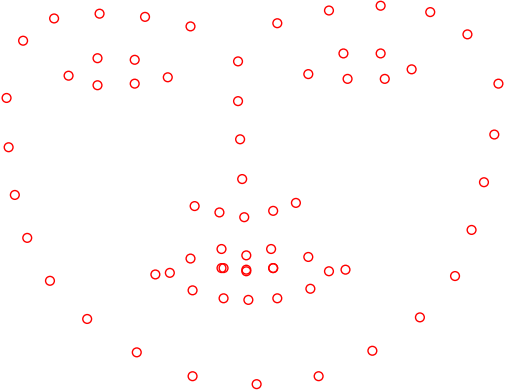} 
    &
    \includegraphics[width=0.22\textwidth]{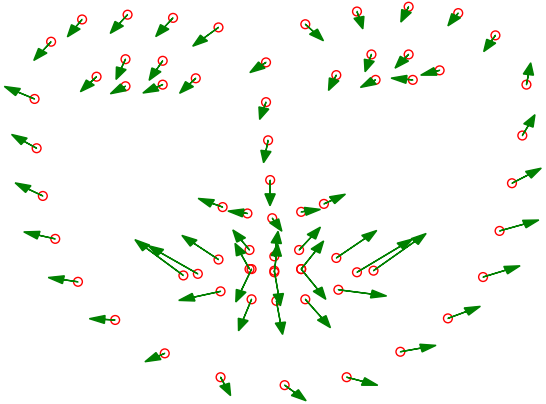}&
    \includegraphics[width=0.22\textwidth]{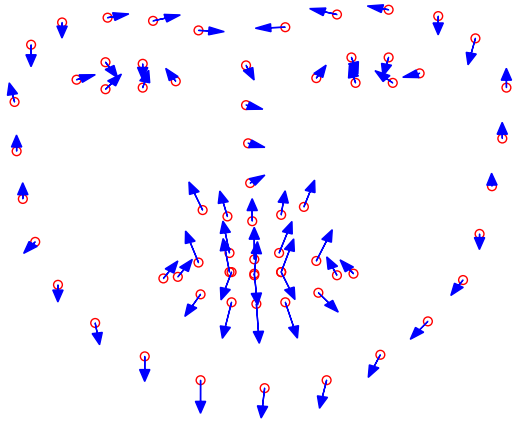} &
 \includegraphics[width=0.22\textwidth]{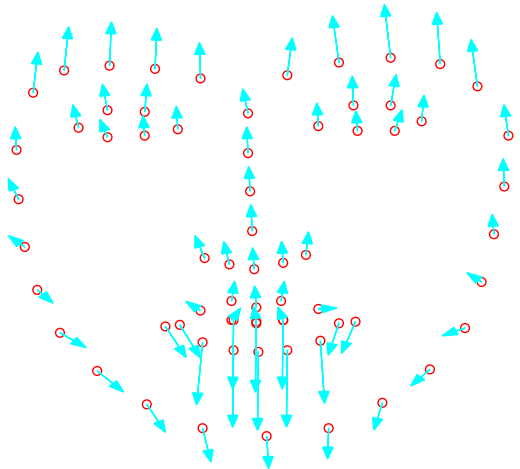}
 \end{tabular}  
    \caption{\label{fig:emotion-vecs}Average 2D shapes of training images: for neutral class 2D shape in the form of 68 detected facial
salient points; for smile class 2D shape in the form of 68 detected facial salient points with vectors from neutral to
smile; for angry class 2D shape in the form of 68 detected facial salient points with vectors from neutral to angry;
 for surprised class 2D shape in the form of 68 detected facial salient points with vectors from neutral to surprised.}
\end{figure}

In Fig.\ref{fig:emotion-vecs} we see that the major information of the facial muscles movements is located at the eyes, eye brows, mouth and jaw. Therefore action units are assigned to this facial parts. The action units having an impact on visible emotion modeling are  given in the table Tab.\ref{tab:aunits}. The particular face expression is always a linear combination of many action units:
\begin{itemize}
\item smile: $1+4+12+20+10+27+25$,
\item anger: $16+4+9+27+10$,
\item surprise: $5+7+1+2+4+26+27+25+10$.
\end{itemize}

\begin{small}
\begin{table}[htb]
\centering
\caption{\label{tab:aunits}Action units considered for muscle movements for visible emotions.}
\begin{tabular}{|l|l|l|l|}
\hline
Indexing & Descrption          & Indexing & Indexing             \\ \hline
1        & Inner brow raiser   & 5        & Upper lid raiser     \\ \hline
2        & Outer brow raiser   & 7        & Lid tightener        \\ \hline
4        & Brow lowerer        & 9        & Nose wrinkler        \\ \hline
26       & Jaw drop            & 12       & Lip corner puller    \\ \hline
25       & Lips part           & 10       & Upper lip raiser     \\ \hline
27       & Mouth stretch       & 15       & Lip corner depressor \\ \hline
16       & Lower lip depressor & 20       & Lip stretcher        \\ \hline
\end{tabular}
\end{table}
\end{small}


Since in the identification of action units parameters we can establish the correspondence of only 37 facial landmarks to Candide head points which have a significant impact on facial expressions, we select only eight action units to be identified:
jaw drop (AU 26/27), brow lower (AU 4), lip corner depressor (AU 13/15), upper lip raiser (AU 10), lip stretcher (AU 20), lid tightener (AU 7), noise wrinkler (AU 9), eye closed (AU 42/43/44/45).

\begin{figure}[htb]
\begin{center}
\begin{tabular}{c|c}
\includegraphics[width=0.35\textwidth]{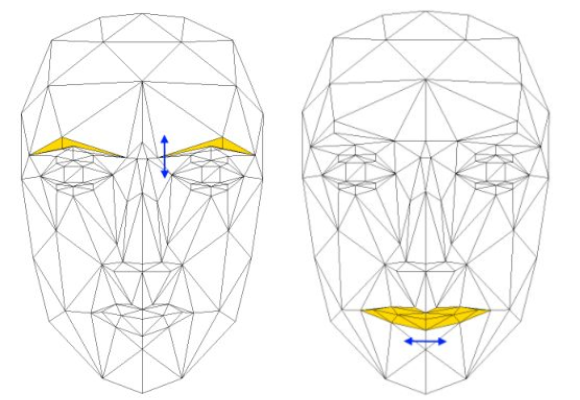}
&
\includegraphics[width=0.35\textwidth]{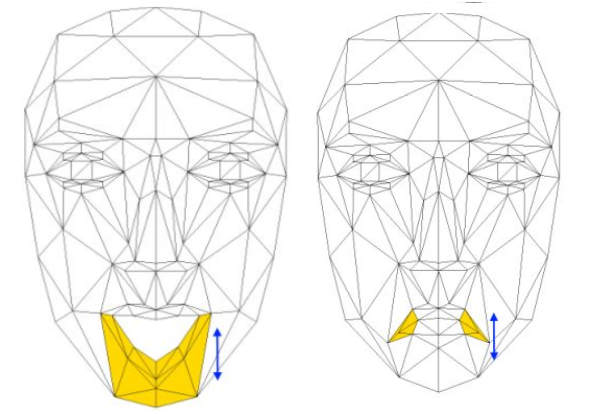}
\end{tabular}
\caption{\label{fig:suau-examples}Examples: on the left the distorted shape units (SU 1, SU 6) and on the right the moved action units (AU 26/27, AU 13/15).}
\end{center}
\end{figure}

While the action units approximate the muscle movements at visible expressions, the shape units indicate the individual (personal) differences for the location and size of facial components. By registering the shape units, we could distort the original mean Candide-3 model into individualized form \cite{su-febriana-skarbek}, thus the action units parameters, we identify, suffer much less influences of their personal information. In the personalization process we select most trusted coefficients out of $15$ shape units including the following seven elements of FACS: eye brows vertical position (SU 1), eyes vertical position (SU 2), eyes width (SU 3), eyes separation distance (SU 4), mouth vertical position (SU 5), mouth width (SU 6), eyes vertical differences (SU 7). Fig.\ref{fig:suau-examples} illustrates the selected shape and the action units.

\subsubsection{Candide modeling  from images -- theory}

The dynamic 3D modeling of human head is represented by the parameters of transformation applied to Candide model. The global motion model has the following form: 
\begin{equation}
P^g_i(\tau) \eqd
\left[
\begin{array}{c}
X^g_i(\tau)\\Y^g_i(\tau)\\Z^g_i(\tau) 
\end{array}
\right]
=
s^g(\tau)R^g(\tau)
\left(
\left[
\begin{array}{c}
X_i^c\\Y_i^c\\Z_i^c 
\end{array}
\right]+
\sum_{d\in[D]:i\in I_d}\alpha_d(\tau)a^d_i
\right)
+t(\tau),\ i\in[G]
\end{equation}
where $i$ is the index of the point in Candide-3 model with $G$ points of global estimation; $I_d$ is the index set of points for the deformation $d \in [D]$, where $D \in G$ is a list of indexes for the selected shape units to be used for personalization (individualization); $a^d_i \in R^3$ is the unit deformation vector being the column of the deformation matrix $A^c_d$ which is assigned to the point $i$ at the deformation $d$, $A^c_d \in R^{3\times |I_d |}$; the notation $d \in [D] : i \in I_d$ selects for the summation only those deformations $d$ which refer to the point $i$.

\begin{figure}[htb]
\centering
\begin{tabular}{cc}
\includegraphics[width=0.4\textwidth]{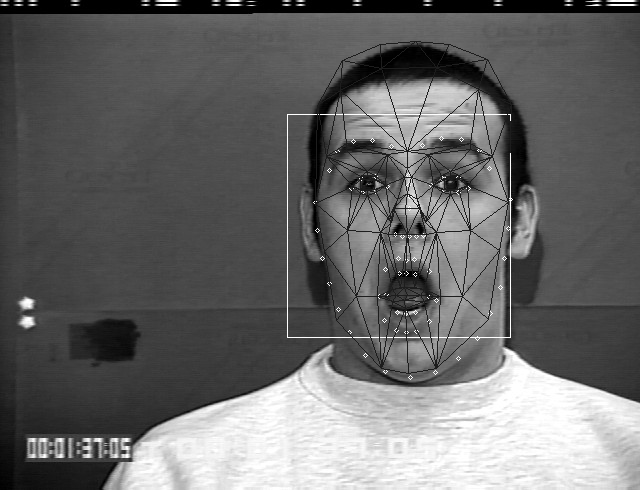}
&
\includegraphics[width=0.4\textwidth]{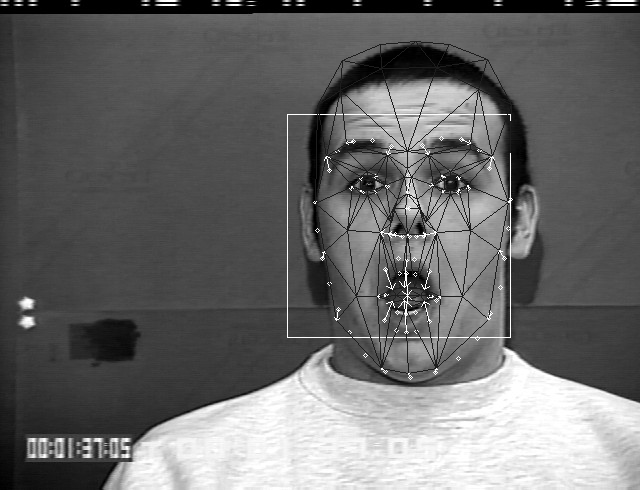}
\end{tabular}
\caption{3D modeling for global estimation and personalization -- projection error shown on the right.}
\end{figure}

Having the global motion parameters and local deformation for personalization, we can relate the local motion of action units with globally moved Candide points (vertexes):
\begin{equation}
P_i(\tau) \eqd
\left[
\begin{array}{c}
X_i(\tau)\\Y_i(\tau)\\Z_i(\tau) 
\end{array}
\right]
=
\left[
\begin{array}{c}
X^g_i(\tau)\\Y^g_i(\tau)\\Z^g_i(\tau) 
\end{array}
\right]
+
s^g(\tau)R^g(\tau)
\left(
\sum_{f\in[F]:i\in I_f}\alpha_f(\tau)a^f_i\right),\ i\in[H]
\end{equation}

where $H$ stands for all $37$ 3D points corresponding to landmarks. The scale parameter and the rotation matrix are the same as those used in global estimation. $I_f$ is the index set of points for the local animation $f \in [F]$, where $F$ is the set of action units while $a^f_i \in R^3$ is the unit deformation vector.

\begin{figure}[htb]
\centering
\begin{tabular}{cc}
\includegraphics[width=0.4\textwidth]{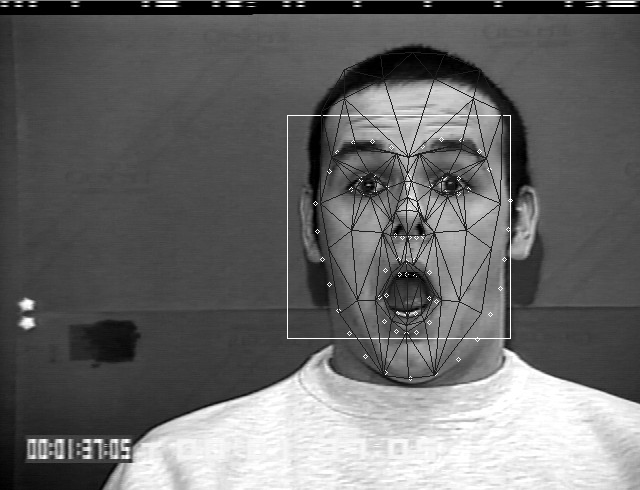}
&
\includegraphics[width=0.4\textwidth]{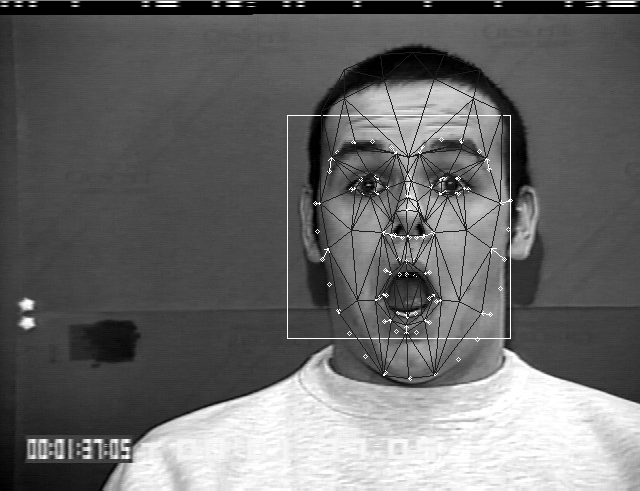}
\end{tabular}
\caption{3D modeling with action units after global estimation and personalization -- projection error shown on the right.}
\end{figure}
The main function of optimization package is to identify the transformation parameters (local deformations for action units and shape units, global scaling, rotation, and translation) of Candide-3 model onto the current face model. To this goal:
\begin{enumerate}
\item Core 3D points for global motion are selected: $J.$
\item Points for global estimation and individualization are selected from core 3D points: $J_g\in J$
\item Indexes of deformation points for shape units are joined to core points: $J_{gd} \eqd \ds\bigcup_{d\in[D]}I_d\cup J_g$
\item Active 2D points $J_s^2$ of facial salient points FP68 having corresponding points in $J_{gd}$ core and deformation points, are selected. 
  \item Active 3D points are specified as those points of $J_{gd}$ which correspond to active 2D points: $J_s^3.$
  \item Number of active points is registered: $N_s=|J_s^2|=|J_s^3|.$
  \item The centroid for Candide model is computed: $\bar{P}^g=
  \left[
\begin{array}{c}
\bar{X}^g\\\bar{Y}^g\\ \bar{Z}^g
\end{array}
\right]
  \eqd\ds
  \frac{1}{N_s}\sum_{i\in J_s^3}
    \left[
\begin{array}{c}
X_i^g\\Y_i^g\\Z_i^g
\end{array}
\right]=
  \frac{1}{N_s}\sum_{i\in J_s^3}P_j^g.$
\end{enumerate}
For the current FP68 shape $p_j(\tau)\inv{2}, j\in J_s^2$, the initial values of motion parameters with respect to Candide-3 shape $P_i^c,$ $i\in J_s^3,$ are found using general formulas \eqref{eq:aff-init}:
\begin{enumerate}
 \item Distortion coefficients and rotation: \begin{equation}
\alpha_d=0,d\in[D],\ R=I_3
\end{equation}
\item Scaling $s$:
\begin{equation}
\begin{array}{c}
s =\ds\arg\min_s\left[\sum_{i\in J_s^3}(x_{j(i)}'-s(X_i^g)')^2+(y_{j(i)}'-s(Y_i^g)')^2\right] \lra
\\[15pt]
 s = \ds\frac{\ds\sum_i\left[x_{j(i)}'(X_i^g)'+y_{j(i)}'(Y_i^g)'\right]}
{\ds\sum_i\left[(X_i^g)'(X_i^g)'+(Y_i^g)'(Y_i^g)'\right]}
\end{array}
\end{equation}
where the 2D/3D centered shapes are defined as follows:
\begin{equation}
\left[
\begin{array}{c}
\bar{x}\\\bar{y}
\end{array}
\right]
\eqd
\ds\frac{1}{N_s}\sum_{j\in J_s^2}
\left[
\begin{array}{c}
x_j\\y_j
\end{array}
\right] \lra 
\left[
\begin{array}{c}
x_j'\\y_j'
\end{array}
\right]
=
\left[
\begin{array}{c}
x_j-\bar{x}\\y_j-\bar{y}
\end{array}
\right]
,
\ \ \ 
\left[
\begin{array}{c}
(X_i^g)'\\(Y_i^g)'
\end{array}
\right]
\eqd
\left[
\begin{array}{c}
X_i^g-\bar{X}^g\\Y_i^g-\bar{Y}^g
\end{array}
\right]
\end{equation}
\item Translation $t:$
\begin{equation}
\left[
\begin{array}{c}
x_{j(i)}\\y_{j(i)}
\end{array}
\right]
\simeq
s\left[
\begin{array}{c}
X_i^c\\Y_i^c
\end{array}
\right]+t,\ \ i\in J_s^3 
\lra
t = \left[
\begin{array}{c}
\bar{x}\\\bar{y}
\end{array}
\right]-s \left[
\begin{array}{c}
\bar{X}^g\\\bar{Y}^g
\end{array}
\right]
\end{equation}
\end{enumerate}
Error function is defined:
\begin{equation}\label{eq:error}
E_{\tau}(s,w,t,a) = \sum_{i\in J_a^3} \left\|P_{i}(\tau)\left|_{xy}\right.-p_{j(i)}(\tau)\right\|^2
\end{equation}
where $s\in\bb{R}$ -- scaling parameter; $w\inv{3}$ -- the vector representation of the rotation matrix (see the inverse Rodrigues formulas below \eqref{eq:inv-rodrig}); $t\inv{2}$ --  the translation vector in the $xy$ plane; $a\inv{D}$ -- parameters of local deformations; $j(i) = j$ such that $J_s^3[k] = i \lra J_s^2[k] = j,$ i.e. it is the active index of 2D point corresponding to the active index of 3D point; $|_{xy}$ -- denotes the orthographic projection onto $xy$ plane.

LMM (Levenberg Marquardt Method) optimization procedure is performed for the error function $E(s,w,t,a)$ defined by equation \eqref{eq:error}with initialization described above.

The function to compute the orthographic projection uses the current transformation parameters. The rotation is represented by 3D vector $w\inv{3}$ representing the rotation angle $\alpha$ in radians $\alpha \eqd \|w\|,$ and the rotation axis $u\eqd\frac{w}{\alpha}.$ The rotation matrix $R$ is found from the Rodrigues formula. 

Namely, let $R$ be the rotation matrix for rotation axis $u$ and rotation angle $\alpha$. If $\tp{x}u=0$ then $Rx=\cos\alpha\cdot x+\sin\alpha\cdot(u\times x),$ otherwise $Rx=R(\tp{u}xu+x-\tp{u}xu)=$ $\tp{u}xRu+R(x-\tp{u}xu)=$ $\tp{u}xu+\cos\alpha(x-\tp{u}xu)+\sin\alpha(u\times(x-\tp{u}xu)).$ Hence $Rx = u\tp{u}x+\cos\alpha\cdot x-$ $\cos\alpha\cdot u\tp{u}x+$ $\sin\alpha(u\times x)=$ $\cos\alpha\cdot x+(1-\cos\alpha) u\tp{u}x+\sin\alpha(u\times x).$ However, 
\begin{equation}
u\times x=\left[
\begin{array}{ccc}
0 & -u_z & u_y\\
u_z & 0 & -u_x\\
-u_y & u_x & 0
\end{array}
\right]x
\lra
R = \cos\alpha\cdot I_3+(1-\cos\alpha)u\tp{u}+\sin\alpha
\left[
\begin{array}{ccc}
0 & -u_z & u_y\\
u_z & 0 & -u_x\\
-u_y & u_x & 0
\end{array}
\right]
\end{equation}

Note that the rotation angle $\alpha$ and the rotation axis $u$ can be recovered from the rotation matrix by the inverse Rodrigues formulas. They follow directly from the linearity of trace and transposition operations for matrices.
\begin{equation}\label{eq:inv-rodrig}
tr{R} = 2\cos\alpha+1,\ \ \ R-\tp{R} = 2\sin\alpha
\left[
\begin{array}{ccc}
0 & -u_z & u_y\\
u_z & 0 & -u_x\\
-u_y & u_x & 0
\end{array}
\right]
\end{equation}

\subsubsection{Affine equations for affine motion initialization}

For orthographic viewing, the affine motion of 3D cloud of points $A_{XYZ}$ without rotation can be considered as the affine motion of 2D model $A_{xy}$. The we scale by $\alpha$ and translate by $[\gamma_x,\gamma_y]$ in order to nearest to our target cloud of points $B_{xy}$. In our case $A_{xy}$ and $B_{xy}$ are Candide model and FP68 landmarks, both restricted to still points only. Having the word {\it nearest} expressed by the least square error we get the following set of equations.

Given four vectors $x_a,x_b,y_a,y_b\inv{n}$, we are looking for three parameters $\alpha,\gamma_x,\gamma_y\in\bb{R}$  such that
\begin{equation}
\left\{
\begin{array}{rcl}
\alpha x_a+\gamma_x1_n &\simeq & x_b\\
\alpha y_a+\gamma_y1_n & \simeq & y_b
\end{array}
\right.
\end{equation}
Then the centering trick leads to the solution:
\begin{equation}\label{eq:aff-init}
\begin{array}{l}
\alpha x_a+\gamma_x1_n\simeq x_b\lra \alpha\bar{x}_a+\gamma_x=\bar{x}_b \lra \alpha x_a'\simeq x_b',\ \gamma_x=\bar{x}_b-\alpha\bar{x}_a\\[5pt]
\alpha y_a+\gamma_y1_n\simeq y_b\lra \alpha\bar{y}_a+\gamma_y=\bar{y}_b \lra \alpha y_a'\simeq y_b',\ \gamma_y=\bar{y}_b-\alpha\bar{y}_a\\[5pt]
\alpha^{\ast} = \arg\min_{\alpha}\left[\|\alpha x_a'-x_b'\|^2+\|\alpha y_a'-y_b'\|^2\right] = 
\frac{\tp{x'}_ax_b'+\tp{y'}_ay_b'}{\|x_a'\|^2+\|y_a'\|^2} 
\end{array}
\end{equation}

\subsubsection{Candide model personalization and local motion normalization}

Candide model personalization is performed using data from neutral expression. From a small number of image frames we get the second order statistics for each shape unit coefficients. Then Gaussian approximation can be used to define the coefficient trust or distrust measure. 
Sorting by a distrust $D(\mu,\sigma)$ of our statistics is based on the cumulative probability distribution in favor of zero value, i.e. we compute the probability of those real values $x$  which are closer to zero than to the mean value:
\begin{equation}
\begin{array}{l}
\ds\mu\leq 0 \lra D(\mu,\sigma) \eqd \int_{\frac{\mu}{2}}^{\infty}G(x;\mu,\sigma)dx = 1-F(x;\mu,\sigma)\\[15pt]
\ds\mu\geq 0 \lra D(\mu,\sigma) \eqd \int_{-\infty}^{\frac{\mu}{2}}G(x;\mu,\sigma)dx = F(x;\mu,\sigma)
\end{array}
\end{equation}
where $F(x;\mu,\sigma)$ is the cumulative probability distribution of the Gaussian $G(x;\mu,\sigma).$

\begin{figure}[htb]
\begin{center}
\begin{tabular}{c}
\includegraphics[width=1.0\linewidth]{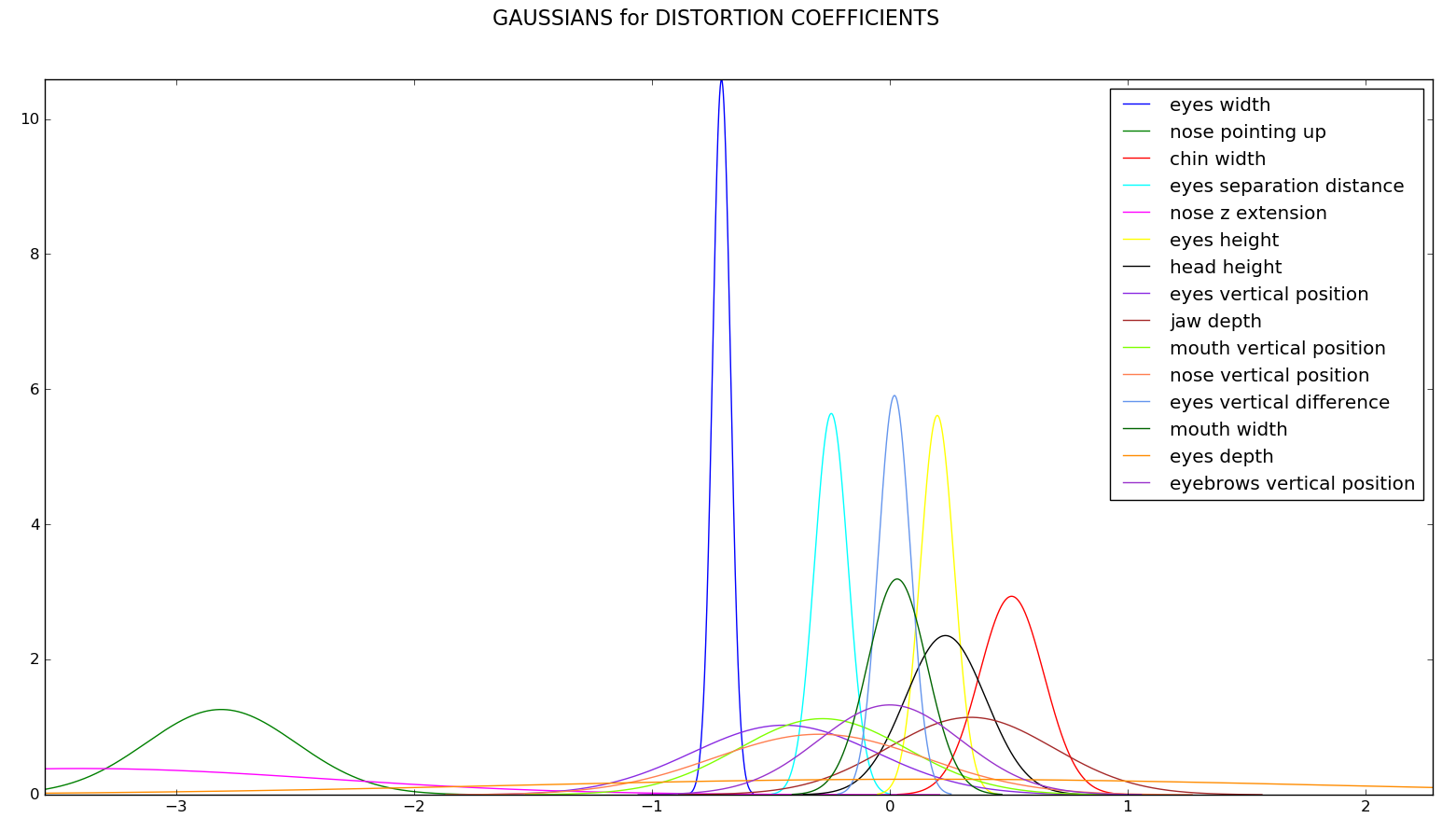}
\end{tabular}
\end{center}
\vspace*{-6mm}
\caption{\label{fig:distortion-Gauss}Gaussian distributions for $n_d=15$ deformation coefficient to fit Candide model to one of author's.}
\end{figure}

In Tab.\ref{tab:distortion-coefficients} the shape coefficients are sorted by the distrust measure. The data is acquired for the person with photo given in Fig.\ref{fig:detect-ws}. All Gaussians characterizing personal face local proportions are drawn in Fig.\ref{fig:distortion-Gauss}.

\begin{footnotesize}
\begin{table}[htb]
\caption{\label{tab:distortion-coefficients}Statistics for $n_d=15$ deformation coefficient to fit Candide model to one of authors. There are seven highly trusted deformations with one additional found below the threshold of distrust. We fix threshold level to $0.25$\ (the single line horizontal rule).}
\vspace*{-2mm}
\[
\begin{array}{rrrrl}
  \text{Distrust} &  \text{Mean} &  \text{Variance} &
  \text{Sigma}    &    Coefficient\ name\\
\hline\hline
     0.000  &  -0.707  &  0.001  &   0.038  & eyes\ width\\
     0.000  &  -2.810  &  0.100  &   0.316  & nose\ pointing\ up\\
     0.030  &   0.513  &  0.018  &   0.136  & chin\ width\\
     0.041  &  -0.245  &  0.005  &   0.071  & eyes\ separation\ distance\\
     0.049  &  -3.398  &  1.051  &   1.025  & nose\ z\ extension\\
     0.080  &   0.200  &  0.005  &   0.071  & eyes\ height\\
     0.244  &   0.234  &  0.029  &   0.169  & head\ height\\
 \hline
     0.284  &  -0.443  &  0.150  &   0.387  & eyes\ vertical\ position\\
     0.312  &   0.340  &  0.121  &   0.348  & jaw\ depth\\
     0.345  &  -0.282  &  0.125  &   0.354  & mouth\ vertical\ position\\
     0.370  &  -0.295  &  0.198  &   0.445  & nose\ vertical\ position\\
     0.440  &   0.020  &   0.005 &    0.068 & eyes\ vertical\ difference\\
     0.450  &   0.031  &   0.016 &    0.125 & mouth\ width\\
     0.483  &   0.144  &   2.991 &    1.729 & eyes\ depth\\
     0.500  &   0.001  &   0.090 &    0.300 & eyebrows\ vertical\ position\\
\hline\hline 
\end{array}
\]
\end{table}
\end{footnotesize}

\paragraph{Normalization for the action units feature vectors\\}

The normalization  of action units feature vectors changes the statistics of animation coefficients to zero mean and unit variance.

Considering that the motion vectors which are defined for each action unit do not share the  same length it is clear that some facial component's coefficients get large values while others are less significant. 
\begin{equation}
x_{norm}=\frac{x-\mu}{\sigma}
\end{equation}
where $x_{norm}$ is the normalized feature vectors and $\mu$ is the mean value of each action unit values  and $\sigma$ is the
standard deviation of each action unit values, respectively.

\section{Datasets for training and testing of classifiers}

\begin{figure}[htb]
    \begin{subfigure}{.5\textwidth}
    \centering
    \includegraphics[width=0.65\textwidth]{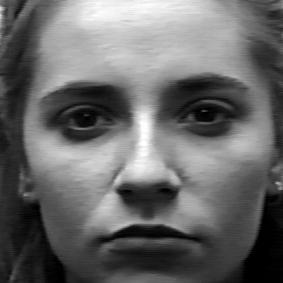}
    \caption{Frontal image from training dataset}
      \label{frontal-img}
    \end{subfigure}
    \begin{subfigure}{.5\textwidth}
    \centering
    \includegraphics[width=0.65\textwidth]{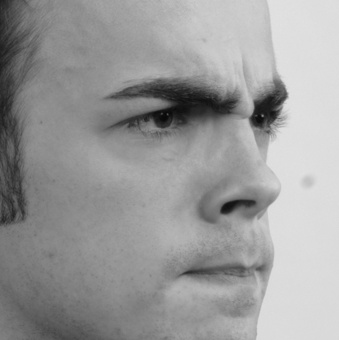}
        \caption{45 degrees pose face image from test dataset}
          \label{45-degree-img}
    \end{subfigure}
\caption{Difference between training pose and testing pose image}
\label{test-train-difference}
\end{figure}

We make experiment to compare classical personalized SVM emotion classifier as prior work. For personalization we choose datasets with neutral face state. Moreover to make experiment more reliable and see how different facial features can affect results we choose the test dataset with also non-frontal face poses (Fig.\ref{45-degree-img}) that are not provided in training data (Fig.\ref{frontal-img}). This selection can show what generalization is  obtained using different classification techniques. For these reason, we don't use open datasets where we can't find neutral faces for personalization as a step of structured-SVM classifier. We use three different databases for training and evaluation. Selected datasets consist of four different emotions for each person. 

The first database we use is Cohn-Kanade dataset established by Lucey et al.  \cite{5543262}, a dataset specified for action units and emotion-specified expression. It includes approximately 2000 video sequences of around 200 people between the age of 18 to 50. It consists of 69\% samples from females and 31\% samples from males. 65\% of them is taken from white people, 15\% of which from black people and 20\% of which from Asian or Latin American. Every video sequences of an individual expression starts from neutral face to the maximized deformation of the face of a certain expression, which provides us the accurate data for action units extraction.

\begin{figure}[htb]
    \begin{subfigure}[b]{.1\linewidth}
    \includegraphics[width=\textwidth]{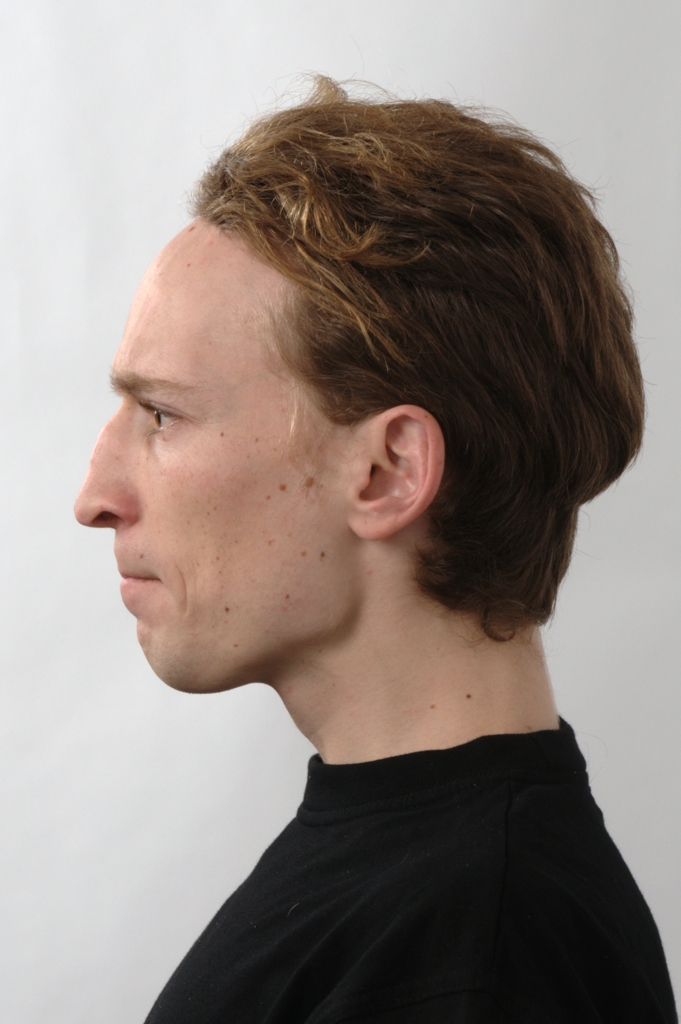}
    \end{subfigure}%
    \hfill
    \begin{subfigure}[b]{.1\linewidth}
    \includegraphics[width=\textwidth]{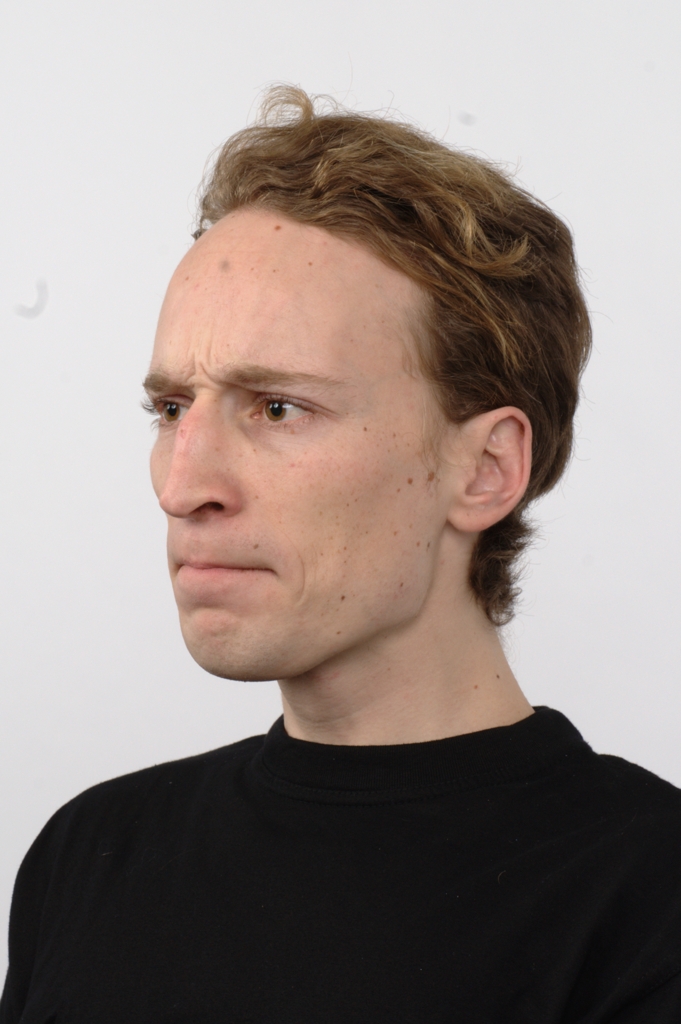}
    \end{subfigure}%
    \hfill
    \begin{subfigure}[b]{.1\linewidth}
    \includegraphics[width=\textwidth]{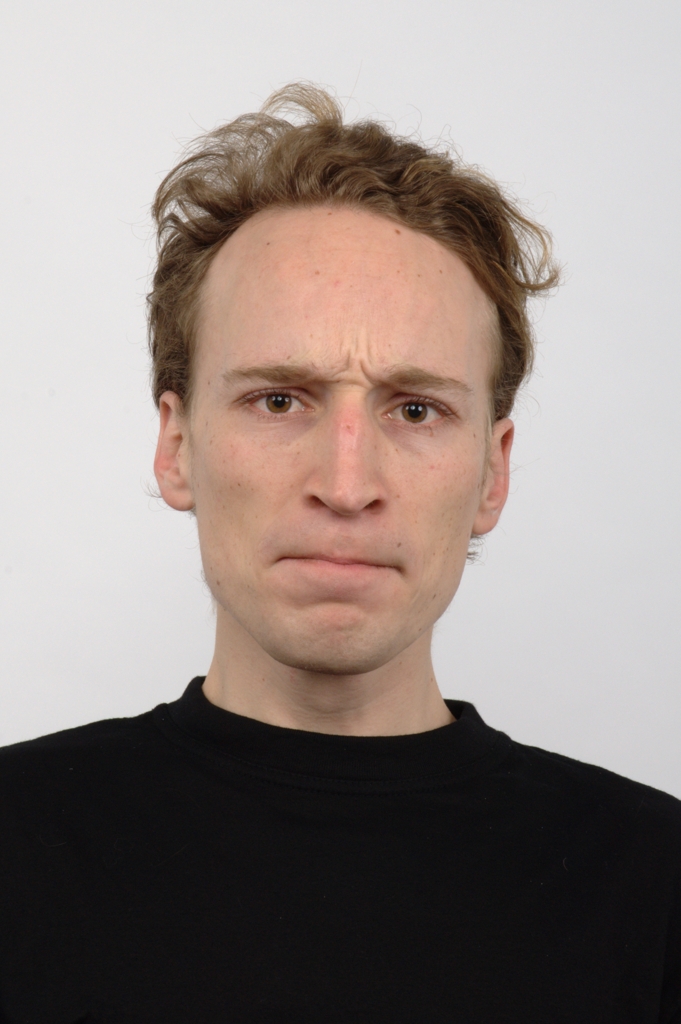}
    \end{subfigure}%
    \hfill
    \begin{subfigure}[b]{.1\textwidth}
    \includegraphics[width=\textwidth]{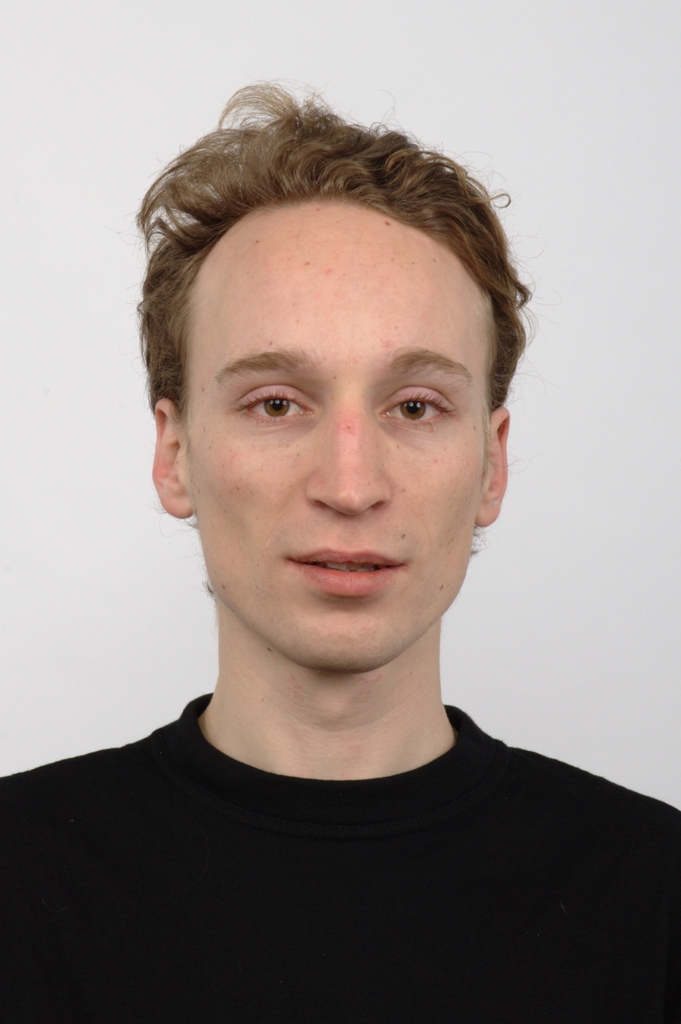}
    \end{subfigure}%
    \hfill
    \begin{subfigure}[b]{.1\textwidth}
    \includegraphics[width=\textwidth]{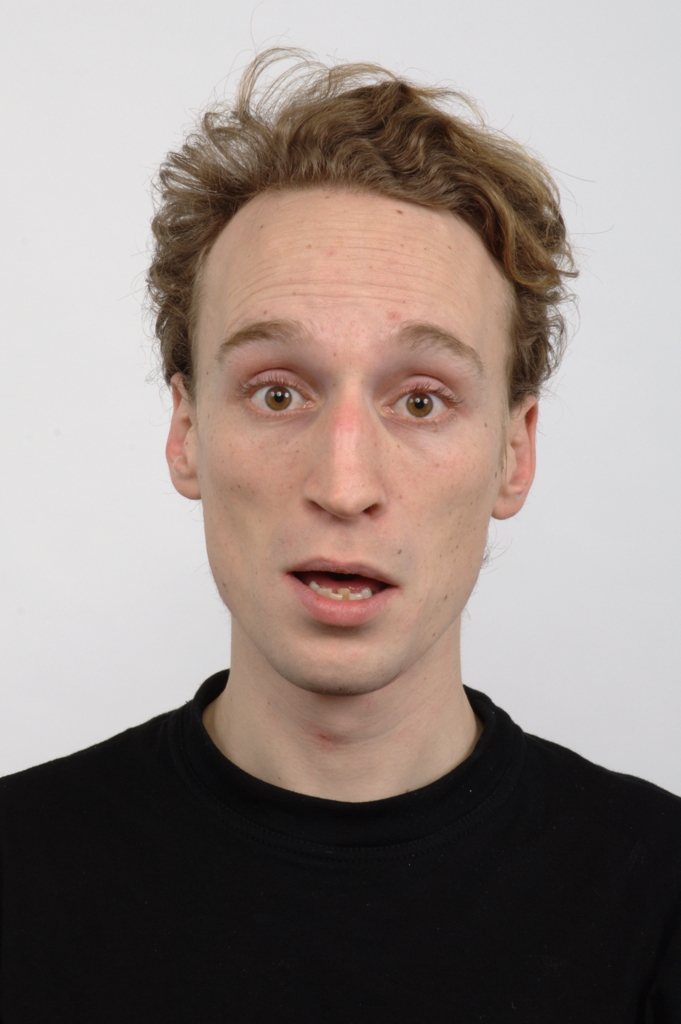}
    \end{subfigure}%
    \hfill
    \begin{subfigure}[b]{.1\textwidth}
    \includegraphics[width=\textwidth]{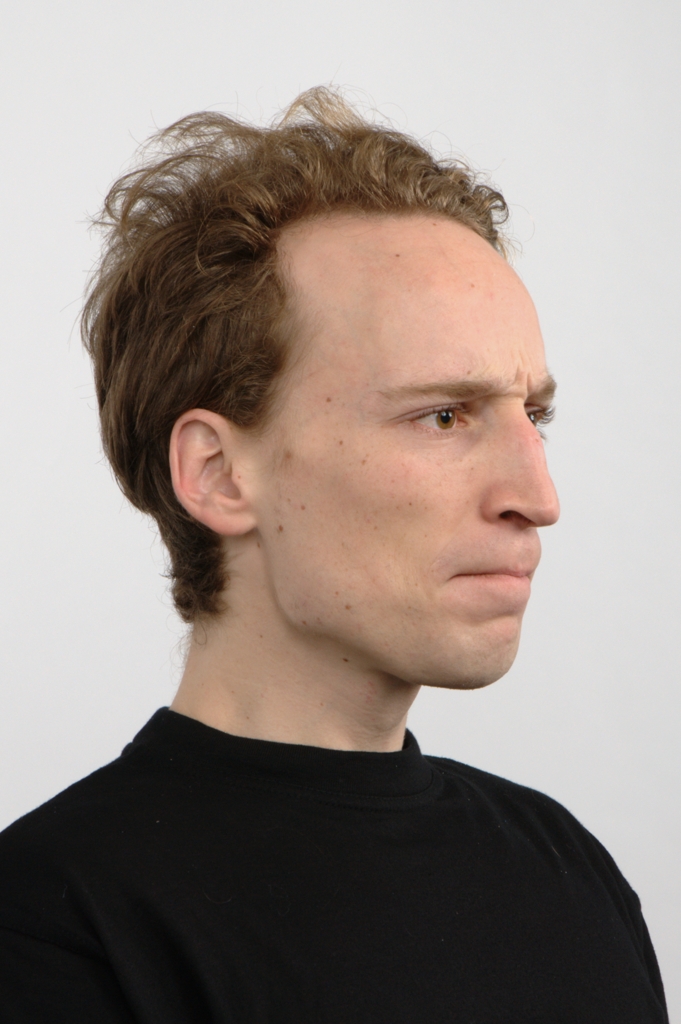}
    \end{subfigure}%
    \hfill
    \begin{subfigure}[b]{.1\textwidth}
    \includegraphics[width=\textwidth]{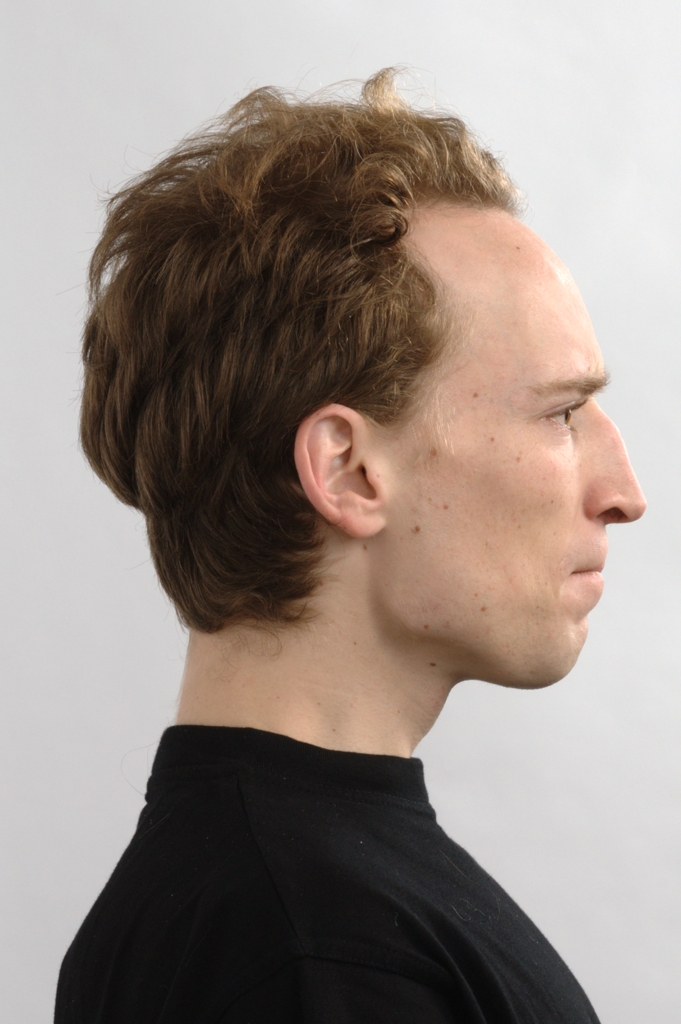}
    \end{subfigure}
\caption{Different face poses (respectively 180, 135, 90, 45, 0 degrees) -- frontal pose is repeated for three different emotions.}
\label{raf-dataset-example}
\end{figure}

The second dataset, The MUG facial expression database developed
by Aifanti et al. \cite{5617662},  consists of image sequences of 86 subjects recorded performing facial expressions. 35 females and 51 males all of Caucasian origin between 20 and 35 years of age. The performing of a special expression are joined with different combination of AUs, e.g., fear can be imitated with mouth closed and only moving the upper part of the face or lips stretched.

\begin{figure}[htb]
    \begin{subfigure}[b]{.23\linewidth}
    \includegraphics[width=\textwidth]{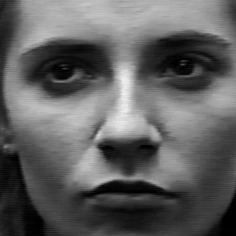}
    \end{subfigure}%
    \hfill
    \begin{subfigure}[b]{.23\linewidth}
    \includegraphics[width=\textwidth]{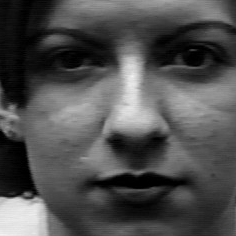}
    \end{subfigure}%
    \hfill
    \begin{subfigure}[b]{.23\linewidth}
    \includegraphics[width=\textwidth]{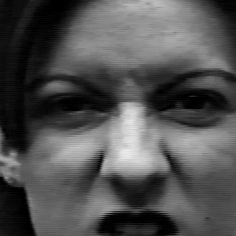}
    \end{subfigure}%
    \hfill
    \begin{subfigure}[b]{.23\textwidth}
    \includegraphics[width=\textwidth]{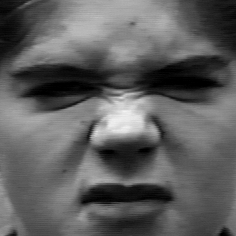}
    \end{subfigure}
    
    \begin{subfigure}[b]{.23\textwidth}
    \includegraphics[width=\textwidth]{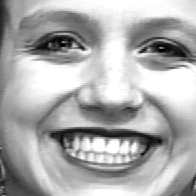}
    \end{subfigure}%
    \hfill
    \begin{subfigure}[b]{.23\textwidth}
    \includegraphics[width=\textwidth]{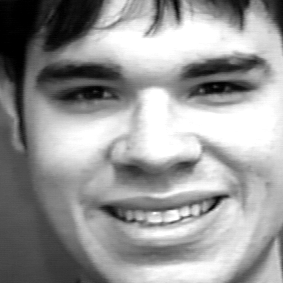}
    \end{subfigure}%
    \hfill
    \begin{subfigure}[b]{.23\textwidth}
    \includegraphics[width=\textwidth]{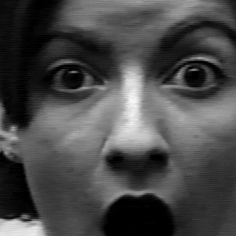}
    \end{subfigure}%
    \hfill
    \begin{subfigure}[b]{.23\textwidth}
    \includegraphics[width=\textwidth]{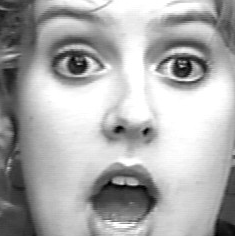}
    \end{subfigure}
\caption{\label{training-cropped-dataset}Cropped training examples for 4 different classes - neutral, angry, smile and surprised}
\end{figure}

The third dataset RaFD which we used for the testing purpose is established by Oliver et al.  \cite{RaFd-dataset}, not just providing us 67 models (including Caucasian males and females, Caucasian children, both boys and girls, and Moroccan Dutch males), but also three samples of an expression of different angles from the same subject, thus the dataset is more challenging for our recognition algorithms to test their performance in practical conditions. 
The RaFD is high quality dataset and contains pictures of eight emotions. Each emotion is shown with three different gaze directions and pictures are taken from five camera angles. We use only three different camera angles (90, 45 and 135 degrees) due to limitations of face detector and facial landmarks extraction, which are not reliable at 180 and 0 degrees (Fig. \ref{raf-dataset-example}).

In total, 6409 image samples were used as training subset, containing 2079 ``neutral'' samples, 2149 ``smile'' samples, 925 ``angry'' samples, and 1256 ``surprised'' samples, respectively from the MUG dataset and Cohn-Kanade 
dataset (Fig.\ref{training-cropped-dataset}) while 2412 samples are equally divided into these four classes for testing purpose from RafD dataset  \cite{RaFd-dataset}.

\section{Classification}

We prepare set of classifiers to recognize emotions using both facial vector features like AU8 and FP68 as well as face grayscale images.

For facial expression  geometric data, the comparison of SVM classifiers and CNN classifiers is performed directly. We train and test them on the same data. 

\subsection{AU8 and FP68 classifiers}


The Support Vector Machine (SVM) finds in high dimensional space, the hyper plane separating the feature vectors with the maximum margin. SVM classifier has proven to be accurate in image processing area and machine learning area in case of limited number of feature samples.

For our experiments we use two different implementations of SVM: structured SVM (SSVM) and SVM using one-against-one decision function with polynomial kernel (SVMpoly). 

To compare SVM and CNN classification techniques, sets of simple neural networks adapted to classify emotions were also created with input AU and FP68 facial features. 

We present the neural architecture in symbolic notation for tensor neural networks with defined BNF grammars in the paper \cite{skarbek-stnl}.

For AU8, where the architecture options are limited due to the low dimensionality of the input data, few kinds of two-tier neural networks have been checked. We observed that enlarging those networks and addition of nonlinear   activations lead to model over-fitting.
\begin{center}
\doublebox{
\begin{tabular}{l}
\xin{a}{1}{au8}\xdense{}{8}{}{}{}\xdense{}{4}{}{}{}\\
\xbound{AU8}{}{
\begin{array}{l}
au8 := 8_a;\ optima := [loss, AdaM, SoftMax]
\end{array}
}
\end{tabular}
}
\end{center}

For FP-68, an architecture was established for the best results using $8$-fold validating process. 
Deepening the neural network, modifications of the activation function, changing tensor dimensions do not improve results neither for training nor for  testing data. However, the dropout regularization technique applied with probability $80\%$ prevents over-fitting.

Multilayer perceptrons are trained with Adam optimizer with starting learning rate 0.001 being reduced by two when validation metric stops improving.
\begin{center}
\doublebox{
\begin{tabular}{l}
\xin{a}{1}{fp68}\xdense{}{16}{}{}{br}\xdrop{80}\xdense{}{4}{}{}{}\\[5pt]
\xbound{FP68}{}{
\begin{array}{l}
fp68 := 136_a;\ optima := [loss, AdaM, SoftMax]
\end{array}
}
\end{tabular}
}
\end{center}


\subsection{Image based emotion classifiers}

We train neural networks on grayscale images for 4-class emotion recognition problem. We think that trained features on small training dataset can be at least good as the features determined by analytic methods. 

Testing and training data require common face detector to crop facial image. From {\tt dlib} library  \cite{king-dlib} we choose CNN/MMOD neural face detector which is more robust to different face poses than HOG/MMOD face detector.

\subsubsection{Image augmentation}

\begin{figure}[htb]

    \begin{subfigure}[b]{.12\linewidth}
    \includegraphics[width=\textwidth]{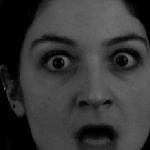}
    \end{subfigure}%
    \hfill
    \begin{subfigure}[b]{.12\linewidth}
    \includegraphics[width=\textwidth]{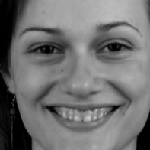}
    \end{subfigure}%
    \hfill\begin{subfigure}[b]{.12\linewidth}
    \includegraphics[width=\textwidth]{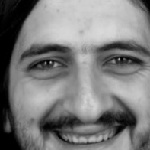}
    \end{subfigure}%
    \hfill\begin{subfigure}[b]{.12\linewidth}
    \includegraphics[width=\textwidth]{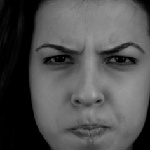}
    \end{subfigure}%
    \hfill\begin{subfigure}[b]{.12\linewidth}
    \includegraphics[width=\textwidth]{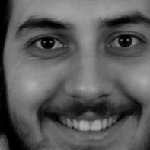}
    \end{subfigure}%
    \hfill\begin{subfigure}[b]{.12\linewidth}
    \includegraphics[width=\textwidth]{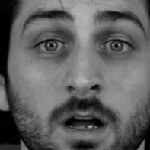}
    \end{subfigure}%
    \hfill\begin{subfigure}[b]{.12\linewidth}
    \includegraphics[width=\textwidth]{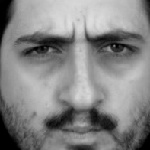}
    \end{subfigure}%
    \hfill\begin{subfigure}[b]{.12\linewidth}
    \includegraphics[width=\textwidth]{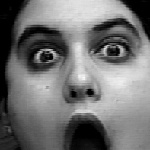}
    \end{subfigure}
    
    \begin{subfigure}[b]{.12\linewidth}
    \includegraphics[width=\textwidth]{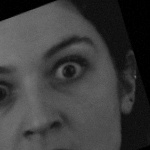}
    \end{subfigure}%
    \hfill\begin{subfigure}[b]{.12\linewidth}
    \includegraphics[width=\textwidth]{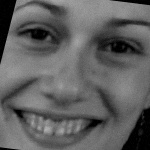}
    \end{subfigure}%
    \hfill\begin{subfigure}[b]{.12\linewidth}
    \includegraphics[width=\textwidth]{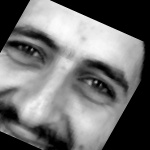}
    \end{subfigure}%
    \hfill\begin{subfigure}[b]{.12\linewidth}
    \includegraphics[width=\textwidth]{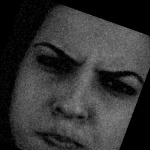}
    \end{subfigure}%
    \hfill\begin{subfigure}[b]{.12\linewidth}
    \includegraphics[width=\textwidth]{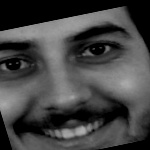}
    \end{subfigure}%
    \hfill\begin{subfigure}[b]{.12\linewidth}
    \includegraphics[width=\textwidth]{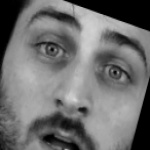}
    \end{subfigure}%
    \hfill\begin{subfigure}[b]{.12\linewidth}
    \includegraphics[width=\textwidth]{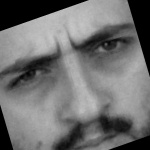}
    \end{subfigure}%
    \hfill\begin{subfigure}[b]{.12\linewidth}
    \includegraphics[width=\textwidth]{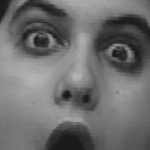}
    \end{subfigure}%
    \hfill
\caption{Cropped faces used for training phase after augmentation procedure}
\label{fig:data-crop-augment}
\end{figure}

The original image training dataset is augmented by performing affine transforms, scalings, cropping, changes of lighting, contrasting, and adding Gaussian noise. The augmentation using {\tt imgaug} library\cite{imgaug}, makes models more robust to changing the pose of the head -- it can be seen in the test set (Fig. \ref{fig:data-crop-augment}).

\begin{figure}[H]
    \begin{subfigure}[b]{.23\linewidth}
    \includegraphics[width=\textwidth]{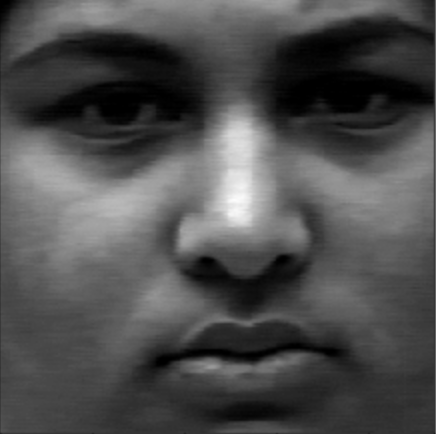}
    \caption{Input image}
    \end{subfigure}%
    \hfill
    \begin{subfigure}[b]{.23\linewidth}
    \includegraphics[width=\textwidth]{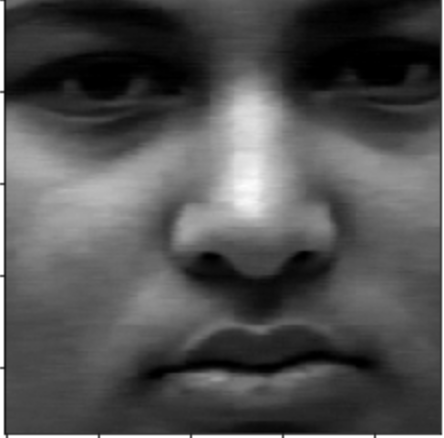}    \caption{Cropping}
    \end{subfigure}%
    \hfill
    \begin{subfigure}[b]{.23\linewidth}
    \includegraphics[width=\textwidth]{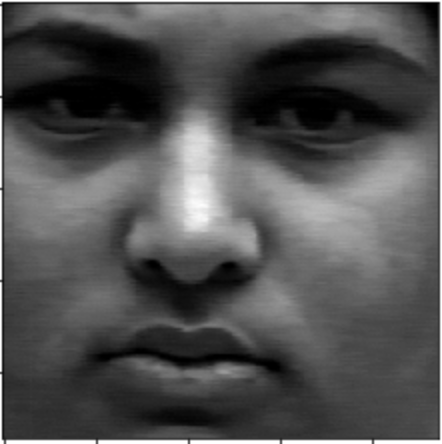}    \caption{Flipping}
    \end{subfigure}%
    \hfill
    \begin{subfigure}[b]{.23\textwidth}
    \includegraphics[width=\textwidth]{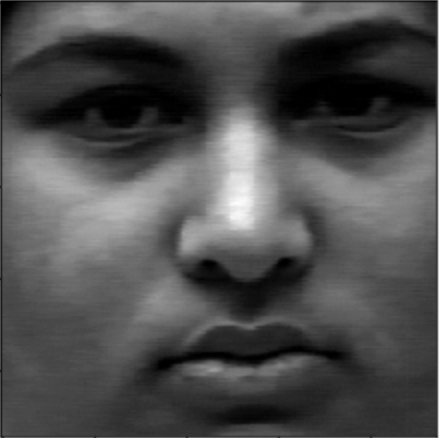}    \caption{Blurring}
    \end{subfigure}
    
    \begin{subfigure}[b]{.23\textwidth}
    \includegraphics[width=\textwidth]{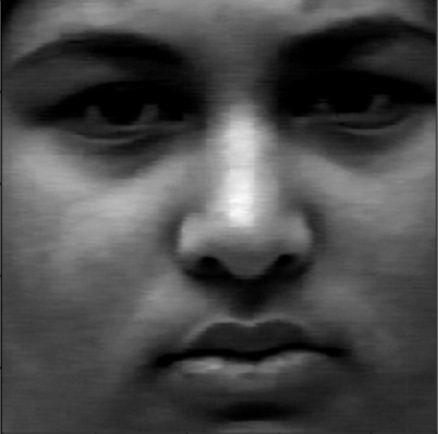}    
    \caption{Contrasting}
    \end{subfigure}%
    \hfill
    \begin{subfigure}[b]{.23\textwidth}
    \includegraphics[width=\textwidth]{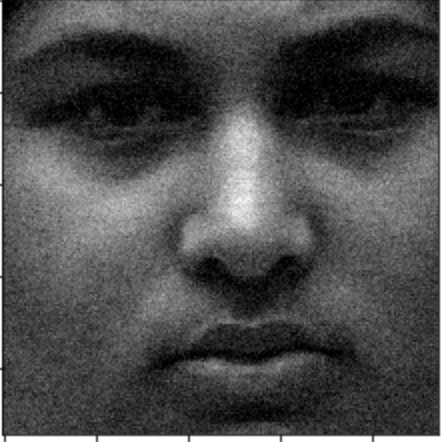}
    \caption{Additive noise}
    \end{subfigure}%
    \hfill
    \begin{subfigure}[b]{.23\textwidth}
    \includegraphics[width=\textwidth]{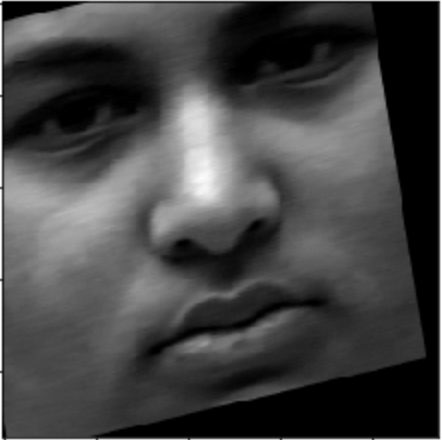}     \caption{Affine transform}
    \end{subfigure}%
    \hfill
    \begin{subfigure}[b]{.23\textwidth}
    \includegraphics[width=\textwidth]{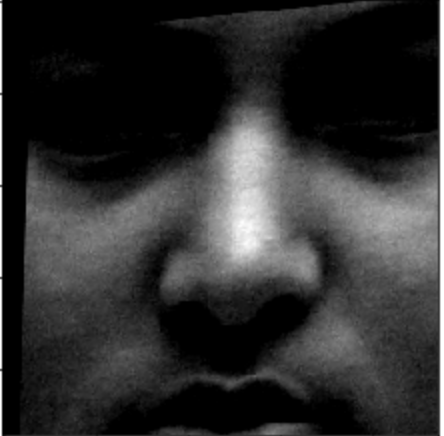}
    \caption{All augmentations}
    \end{subfigure}
\caption{Image augmentation results, final augmentation consists of all operations applied in random order}
\label{fig:original-before-aug}
\end{figure}

We define list of image processing operations on training images as augmentation procedure on image $I_i$ (Fig. \ref{fig:original-before-aug}). Augmentation consists of stochastic set of procedures. Some of them are applied with particular probability. Order of augmentation is also randomized to provide better mixture of training data.

\begin{enumerate}
    \item Vertical axis symmetry is applied with probability $0.5$;
    \item Cropping randomly $0-10\%$ rows and columns of the image;
    \item Gaussian blur $\cl{N}(0,\sigma)$ is randomly applied with probability $0.5$ for $\sigma\in\mathcal{U}(0,0.5);$
    
    \item Contrast normalization\\
    $\alpha \ass \mathcal{U} (0.75,1.5);\quad
    I_i' \ass \max(0,\min(255,\alpha\cdot (I_i - 128) + 128))$
    
    \item Additive Gaussian Noise\\
    $Z_i\ass \mathcal{N}(0,25);\quad 
    I_i' \ass \max(0,\min(255,I_i + Z_i))$
    
    \item Affine transform with random matrix in the uniform pixel coordinates representing the composition of the following basic transformations:
    \begin{itemize}
    \item scaling by $s\in\cl{U}(0.9,1.1),$
    \item translating $t_x\in\cl{U}(-x_{res}/10,x_{res}/10)$, $t_y\in\cl{U}(-y_{res}/10,y_{res}/10),$
    \item rotating by $\theta\in\cl{U}(-25^{\circ},25^{\circ}),$
    \item shearing by $\alpha\in\cl{U}(-8^{\circ},8^{\circ}).$
    \end{itemize}

\end{enumerate}

\subsubsection{Neural networks}

We also test convolutional neural networks on different cropped face image sizes: 150x150, 75x75 and 50x50. This experiment shows what is general impact of resolution on the performance. For each image size the best architecture is chosen (Tab.\ref{results-accuracy}). 
Detailed architecture consists of several convolutional, max pooling, dropout and fully connected layers. 

{\it CNN-1} network is constructed of convolutional layers with batch normalization and non-linear ReLU activation unit. After the last convolution layer the global average pooling is used. The network doesn't contain fully connected layer thanks to the preceding convolution layer which defines four feature maps.
\begin{center}
\doublebox{
\begin{tabular}{l}
\xin{yx}{}{img50}
\xconv{5}{16}{p}{}{b}
\xconv{2_{\sigma}5}{16}{p}{}{br}
\xconv{5}{32}{p}{}{b}
\xconv{2_{\sigma}5}{32}{p}{}{br}
\xconv{3}{64}{p}{}{b}
\xconv{2_{\sigma}3}{64}{p}{}{br}
\xconv{1}{64}{p}{}{b}\\[5pt]
\hspace*{2cm}
\xconv{2_{\sigma}3}{128}{p}{}{br}
\xconv{1}{256}{p}{}{b}
\xconv{2_{\sigma}3}{128}{p}{}{}
\xconv{1}{256}{p}{}{b}
\xconv{2_{\sigma}3}{4}{p}{}{}
\xpool{g}{}{a}{}{}
\\
\xbound{cnn-1}{}{
\begin{array}{l}
img50 := 50_{yx};\ optima := [loss, AdaM, SoftMax]
\end{array}
}
\end{tabular}
}
\end{center}

{\it CNN-2} network is inspired by {\it xception} architecture  \cite{chollet_xception:_2016}. It contains cast adder blocks with depth-wise separable convolution layers. Global average pooling is also used in the same way as in the network {\it CNN-1}.

\begin{center}
\noindent\doublebox{
\begin{tabular}{l}
\xin{yx}{}{img75}
\xconv{3}{8}{p}{}{br}
\xconv{3}{8}{p}{}{br}
\\[5pt]
\hspace*{2cm}
\xresid{
\xconv{2_{\sigma}1}{16}{p}{}{~r}\Big| 
\xconv{3}{16}{ps_d}{}{br}
\xconv{3}{16}{ps_d}{}{b}
\xpool{2_{\sigma}3}{}{m}{}{}
}{}
\xresid{
\xconv{2_{\sigma}1}{32}{p}{}{~r}\Big| 
\xconv{3}{32}{ps_d}{}{br}
\xconv{3}{32}{ps_d}{}{b}
\xpool{2_{\sigma}3}{}{m}{}{}
}{}\\[5pt]
\hspace*{2cm}
\xresid{
\xconv{2_{\sigma}1}{64}{p}{}{~r}\Big| 
\xconv{3}{64}{ps_d}{}{br}
\xconv{3}{64}{ps_d}{}{b}
\xpool{2_{\sigma}3}{}{m}{}{}
}{}
\xresid{
\xconv{2_{\sigma}1}{128}{p}{}{~r}\Big|
\xconv{3}{128}{ps_d}{}{br}
\xconv{3}{128}{ps_d}{}{b}
\xpool{2_{\sigma}3}{}{m}{}{}
}{}\\[5pt]
\hspace*{2cm}
\xconv{3}{4}{p}{}{}
\xpool{g}{}{a}{}{}
\\
\xbound{cnn-2}{}{
\begin{array}{l}
img75 := 75_{yx};\ optima := [loss, AdaM, SoftMax]
\end{array}
}
\end{tabular}
}
\end{center}

The above unstructured form can be simplified by exploiting user defined units\footnote{Note that the notation of the residual block $\langle units\rangle$ is generalized now to multi residual block $\langle units|units|units\rangle$ also known as the cast adder.} (cf. \cite{skarbek-stnl}).

\begin{center}
\noindent\doublebox{
\begin{tabular}{l}
\xunitdef{xcept}{
\xresid{
\xconv{2_{\sigma}1}{1_{\$}}{p}{}{~r}\Big| 
\xconv{3}{1_{\$}}{ps_d}{}{br}
\xconv{3}{1_{\$}}{ps_d}{}{b}
\xpool{2_{\sigma}3}{}{m}{}{}
}{}
}
\xunitinstance{xcept}{1}{16}
\xunitinstance{xcept}{2}{32}
\xunitinstance{xcept}{3}{64}
\xunitinstance{xcept}{4}{128}
\\[10pt]
\xin{yx}{}{img75}
\xconv{3}{8}{p}{}{br}
\xconv{3}{8}{p}{}{br}
\xunit{xcept}{1}{}
\xunit{xcept}{2}{}
\xunit{xcept}{3}{}
\xunit{xcept}{4}{}
\xconv{3}{4}{p}{}{}
\xpool{g}{}{a}{}{}
\\
\xbound{cnn-2}{}{
\begin{array}{l}
img75 := 75_{yx};\ optima := [loss, AdaM, SoftMax]
\end{array}
}
\end{tabular}
}
\end{center}

The above network {\it CNN-2} exhibits comparable results to {\it CNN-1} for testing data. However, its architecture is more complicated what in this case leads to better generalization  measured by the difference between performance for training and testing data:\\ $0.927-0.847>0.865-0.836$ (cf. Tab.\ref{results-accuracy}).

The last network {\it CNN-3} is built of convolutional, max pooling and dense layers with the first followed by dropout layer during the training stage.
\begin{center}
\doublebox{
\begin{tabular}{l}
\xin{yx}{}{img150}
\xconv{3}{32}{}{}{br}
\xpool{2}{}{m}{}{}
\xconv{3}{32}{}{}{br}
\xpool{2}{}{m}{}{}
\xconv{3}{64}{}{}{br}
\xpool{2}{}{m}{}{}
\xconv{3}{64}{}{}{br}
\xpool{2}{}{m}{}{}
\xdense{}{64}{}{}{r}
\xdrop{50}
\xdense{}{4}{}{}{}\\
\xbound{cnn-3}{}{
\begin{array}{l}
img150 := 150_{yx};\ optima := [loss, AdaM, SoftMax]
\end{array}
}
\end{tabular}
}
\end{center}

It is interesting that the architecture of {\it CNN-3} is simpler than architecture of {\it CNN-1}. Apparently, it follows from density of image details for higher image resolution.

\section{Experimental results}

Statistical results compare SVM and DNN using different features.
The statistics computations share the same training and testing samples between SVM and DNN for the discriminative expression features. To analyze the details of the performance and to weight the ups and downs in various aspects for different features and algorithms, we selected Accuracy, 
 Cohen's kappa value and $F_1 score$ as 
 the performance measures. 

\subsection{Accuracy}
The accuracy takes the simple average success rate as the final score, counted by
\[
accuracy = \frac{\text{\it number of correct predictions in all classes}}{\text{\it number of total predictions in all classes}}
\]

Statistics from Tab.\ref{tab:acu-res} indicate that when dealing with the same discriminative features both for AU and FP68, DNN's solution are overwhelmingly more accurate than SVM's solutions. We also prepare cross-validation methodology to test performance of SVM classifiers and select the best. In the Tab. \ref{std-mean-svm} we put standard deviation and mean of accuracy for svm classifiers. Statisitfcs are collected from 30 different experiments. We observe that standard deviation is small for statistics, so each model performance is similar. \\
AU's results are almost $15\%$ more accurate than the pure geometric FP68, with DNN's classification algorithms, it reaches almost the same level as the simple {\it CNN-1}, 50x50 result. 
Having the robust capability of dealing with the RGB images as the input themselves, DNN's results even peak at $87.7\%$ while using the classical features as input gives the lower accuracy at $75.4$\%.

\begin{table}[htb]
\begin{center}
\caption{\label{tab:acu-res}Accuracy results for selected features}
\label{results-accuracy}
\begin{tabular}{|l|l|l|l|l|}
\hline
                         & \multicolumn{2}{c|}{\textbf{Train Data}} & \multicolumn{2}{c|}{\textbf{Test Data}} \\ \hline
\textbf{Vectorized Data} & \textbf{AU}       & \textbf{FP-68}       & \textbf{AU}       & \textbf{FP-68}      \\ \hline
SSVM                     & 0.838             & 0.800                & 0.411             & 0.335               \\ \hline
SVM (poly)               & 0.824             & 0.611                & 0.442             & 0.404               \\ \hline
DNN*                     & 0.830             & 0.642                & 0.754             & 0.611               \\ \hline
\multicolumn{5}{|c|}{\textbf{Images}}                                                                         \\ \hline
{\it CNN-1} 50x50x1            & \multicolumn{2}{l|}{0.838}               & \multicolumn{2}{l|}{0.763}              \\ \hline
{\it CNN-1} 75x75x1            & \multicolumn{2}{l|}{0.927}               & \multicolumn{2}{l|}{0.847}              \\ \hline
{\it CNN-2} 75x75x1            & \multicolumn{2}{l|}{0.865}               & \multicolumn{2}{l|}{0.836}              \\ \hline
{\it CNN-3} 150x150x1          & \multicolumn{2}{l|}{0.932}               & \multicolumn{2}{l|}{0.877}              \\ \hline   
\end{tabular}\\[5pt]
{\footnotesize Note: DNN$^{\ast}$ -- for each input data type there is different architecture.} 
\end{center}
\end{table}

\begin{table}[]
\begin{center}
\caption{Standard deviation and mean of accuracy for SVM classifiers }
\label{std-mean-svm}
\begin{tabular}{|l|l|l|l|l|}
\hline
 & \multicolumn{2}{c|}{\textbf{Train Data}} & \multicolumn{2}{c|}{\textbf{Test Data}} \\ \hline
\multicolumn{5}{|c|}{\textbf{Mean of SR}} \\ \hline
\textbf{Vectorized Data} & AU & FP68 & AU & FP68 \\ \hline
SVM (poly) & 0.799 & 0.605 & 0.426 & 0.388 \\ \hline
SSVM & 0.835 & 0.746 & 0.404 & 0.311 \\ \hline
\multicolumn{5}{|c|}{\textbf{Standard deviation of SR}} \\ \hline
\textbf{Vectorized Data} & AU & FP68 & AU & FP68 \\ \hline
SVM (poly) & 0.011 & 0.003 & 0.008 & 0.008 \\ \hline
SSVM & 0.002 & 0.038 & 0.004 & 0.027 \\ \hline
\end{tabular}\\[5pt]
\end{center}
\end{table}

\subsection{Cohen's Kappa results}

Let $C_{ij}$ is the number of testing examples which belong to the class $i\in[K]$ but they are recognized to be from the class $j\in[K].$
Beside the confusion matrix $C\inm{K}{K}$, the probability $p\inv{K}$ of detection for each class of detector is estimated and Cohen's $\kappa\in\bb{R}$ coefficient is computed. We use the following formulas:
\begin{equation}
\begin{array}{l}
\ds N = \sum_{i\in[K]}\ds\sum_{j\in[K]} C_{ij}\quad\quad\quad
\boxed{p_o \eqd \ds\frac{1}{N}\cdot\sum_{k\in[K]}C_{kk}}\\[20pt]
\ds p_1[i] \eqd \frac{1}{N}\cdot \ds\sum_{j\in[K]} C_{ij},\ i\in[K], \quad 
\ds p_2[j] \eqd \frac{1}{N}\cdot \ds\sum_{i\in[K]} C_{ij},\ j\in[K]\\[10pt]
\ds\boxed{p_e \eqd \sum_{k\in[K]} p_1[k]* p_2[k]} \quad\quad\quad
\boxed{\kappa\eqd \frac{p_o-p_e}{1-p_e}}
\end{array}
\end{equation}

Cohen's kappa value $\kappa$ is a statistical way of measuring the inter-rater agreement (accuracy) for the classes. Instead of only counting the percentage of the correct prediction, Cohen's kappa value takes the possibility of the prediction's occurring by a chance. It means that $p_e$ is the probability of the random agreement and $p_o$ stands for the observed accuracy (agreement).

From the results in Tab.\ref{tab:kappa-res} we observe that for AU input data, the  CNN algorithm  exceeds the SVM solutions in $\kappa$ measure by at least $50\%$ while in for FP68 the Cohen's kappa is higher more than three times. 

For raw image data the advantage of CNN solutions over geometric data changes of $\kappa$ with pixel resolution from $0.01$  to $0.017$. The conclusion about higher generalization of {\it xception} architecture {\it CNN-2} over simpler architecture {\it CNN-1} is valid for both measures: the accuracy and the Cohen's kappa.

\begin{table}[htb]
\centering
\caption{\label{tab:kappa-res}Cohen's kappa results for selected features (SVM, DNN).}
\label{f1-score-results}
\begin{tabular}{|l|l|l|l|l|}
\hline
                         & \multicolumn{2}{c|}{\textbf{Train Data}} & \multicolumn{2}{c|}{\textbf{Test Data}} \\ \hline
\textbf{Vectorized Data} & \textbf{AU}       & \textbf{FP-68}       & \textbf{AU}       & \textbf{FP-68}      \\ \hline
SSVM                     & 0.772             & 0.718                & 0.215             & 0.112               \\ \hline
SVM (poly)               & 0.748             & 0.430                & 0.256             & 0.204               \\ \hline
DNN$^{\ast}$                    & 0.758             & 0.485                & 0.673             & 0.482               \\ \hline
\multicolumn{5}{|c|}{\textbf{Images}}                                                                         \\ \hline
{\it CNN-1} 50x50x1            & \multicolumn{2}{l|}{0.775}               & \multicolumn{2}{l|}{0.684}              \\ \hline
{\it CNN-1} 75x75x1            & \multicolumn{2}{l|}{0.899}               & \multicolumn{2}{l|}{0.798}              \\ \hline
{\it CNN-2} 75x75x1            & \multicolumn{2}{l|}{0.814}               & \multicolumn{2}{l|}{0.782}              \\ \hline
{\it CNN-3} 150x150x1          & \multicolumn{2}{l|}{0.905}               & \multicolumn{2}{l|}{0.836}              \\ \hline
\end{tabular}\\[5pt]
{\footnotesize Note: DNN$^{\ast}$ -- for each input data type there is different architecture.} 
\end{table}

\subsection{Weighted $F_1$ score evaluation}

Performance measure (index) $F_{\beta}$ also known as F-measure, F-score shows the weighted average of precision and recall. 
Using the $\beta\geq 0$ parameter, this measure combines the precision and recall measures into one performance measure:
\begin{equation}\label{eq:F-index}
 F_{\beta}\eqd \ds\frac{1+\beta^2}
{\frac{1}{precision}+\frac{\beta^2}{recall}} \lra 
F_1\eqd 2\cdot\frac{precision\cdot recall}{precision+recall}
\end{equation}
We compute $F_1$ score which reduces F-score to double harmonic average of precision and recall measures. As expected, weighted $F_1$ score again lead to the similar conclusions as we have seen for the accuracy and the Cohen's kappa measures.

\begin{table}[htb]
\centering
\caption{Weighted $F_1$ score results for selected features}
\label{weighted-f1-results}
\begin{tabular}{|l|l|l|l|l|}
\hline
                         & \multicolumn{2}{c|}{\textbf{Train Data}} & \multicolumn{2}{c|}{\textbf{Test Data}} \\ \hline
\textbf{Vectorized Data} & \textbf{AU}       & \textbf{FP-68}       & \textbf{AU}       & \textbf{FP-68}      \\ \hline
SSVM                     & 0.810             & 0.786                & 0.354             & 0.228               \\ \hline
SVM (poly)               & 0.786             & 0.568                & 0.406             & 0.266               \\ \hline
DNN*                     & 0.808             & 0.615                & 0.752             & 0.581               \\ \hline
\multicolumn{5}{|c|}{\textbf{Images}}                                                                         \\ \hline
{\it CNN-1} 50x50x1            & \multicolumn{2}{l|}{0.837}               & \multicolumn{2}{l|}{0.760}              \\ \hline
{\it CNN-1} 75x75x1            & \multicolumn{2}{l|}{0.928}               & \multicolumn{2}{l|}{0.849}              \\ \hline
{\it CNN-2} 75x75x1            & \multicolumn{2}{l|}{0.865}               & \multicolumn{2}{l|}{0.834}              \\ \hline
{\it CNN-3} 150x150x1          & \multicolumn{2}{l|}{0.933}               & \multicolumn{2}{l|}{0.880}              \\ \hline
\end{tabular}\\[5pt]
{\footnotesize Note: DNN$^{\ast}$ -- for each input data type there is different architecture.} 
\end{table}

\section{Conclusions}

\begin{small}
\begin{table}[htb]
\begin{center}
\caption{Neural architectures for {\it emotions from images}.}
\label{tab:all-nets}
\begin{tabular}{c|c|l}
Input & SR[\%] & Symbolic notation for neural architecture\\
\hline
{\it FP68} & 57 & \xin{a}{1}{fp68}\xdense{}{16}{}{}{br}\xdrop{80}\xdense{}{4}{}{}{}\\[5pt]
{\it AU8} & 75 & \xin{a}{1}{au8}\xdense{}{8}{}{}{}\xdense{}{4}{}{}{}\\[5pt]
$50_{xy}$ & 76 & \xin{yx}{}{img}
\xunit{bbr}{1}{}\xunit{bbr}{2}{}\xunit{bbr}{3}{}
\xconv{1}{64}{p}{}{b}
\xconv{2_{\sigma}3}{128}{p}{}{br}
\xconv{1}{256}{p}{}{b}
\xconv{2_{\sigma}3}{128}{p}{}{}
\xconv{1}{256}{p}{}{b}
\xconv{2_{\sigma}3}{4}{p}{}{}
\xpool{g}{}{a}{}{}\\[5pt]
$75_{xy}$ & 83 & \xin{yx}{}{img}
\xconv{3}{8}{p}{}{br}
\xconv{3}{8}{p}{}{br}
\xunit{xcept}{1}{}
\xunit{xcept}{2}{}
\xunit{xcept}{3}{}
\xunit{xcept}{4}{}
\xconv{3}{4}{p}{}{}
\xpool{g}{}{a}{}{}\\[5pt]
$150_{xy}$ & 88 & \xin{yx}{}{img}
\xconv{3}{32}{}{}{br}
\xpool{2}{}{m}{}{}
\xconv{3}{32}{}{}{br}
\xpool{2}{}{m}{}{}
\xconv{3}{64}{}{}{br}
\xpool{2}{}{m}{}{}
\xconv{3}{64}{}{}{br}
\xpool{2}{}{m}{}{}
\xdense{}{64}{}{}{r}
\xdrop{50}
\xdense{}{4}{}{}{}
\end{tabular}\\[5pt]
\end{center}
where the unit {\it bbr} is defined for $50_{xy}$: \xunitdef{bbr}{
\xconv{1_{\$}}{2_{\$}}{p}{}{b}
\xconv{2_{\sigma}1_{\$}}{2_{\$}}{p}{}{br}
}
\xunitinstance{bbr}{1}{5,16}
\xunitinstance{bbr}{2}{5,32}
\xunitinstance{bbr}{3}{3,64}
\end{table}
\end{small}

The experiments regarding to the facial expression classification performance of different features(raw images, FP68 landmarks and action units) and algorithms(SVM and DNN) illustrate that when dealing with each type of those discriminative features, DNN as the classification algorithm shows the most promising results, even when just classifying the eight dimensional data, it holds approximately solid 30\% advantage in accuracy than SVM when the testing samples are much more challenging than the training samples.

Namely, at the challenging conditions when the models are trained for frontal views of human faces while they are tested for arbitrary head poses, for geometric features, the success rate (accuracy)  indicate  nearly triple increase of performance of CNN with respect to SVM classifiers. For raw images, CNN outperforms in accuracy  its best geometric counterpart (AU/CNN) by about 30 percent while the best SVM solutions are inferior nearly four times. For F-score the high advantage of raw/CNN over geometric/CNN and geometric/SVM is observed, as well. 

We conclude also that contrary to CNN based emotion classifiers, the generalization capability wrt human head pose is for SVM based emotion classifiers poor. 

To summarize and compare the neural architectures and their performance we assemble them in Tab.\ref{tab:all-nets} sorting by type of input with the success rate column SR[\%].
 


\bibliographystyle{unsrt}
\bibliography{citations}
\end{document}